\documentclass[manuscript,screen]{acmart}
\usepackage{graphicx}%
\usepackage{multirow}%
\usepackage{amsmath,amsfonts}%
\usepackage{amsthm}%
\usepackage{mathrsfs}%
\usepackage[title]{appendix}%
\usepackage{xcolor}%
\usepackage[most]{tcolorbox}%
\usepackage{textcomp}%
\usepackage{manyfoot}%
\usepackage{booktabs}%
\usepackage{algorithm}%
\usepackage{algorithmicx}%
\usepackage{algpseudocode}%
\usepackage{listings}%
\usepackage[edges]{forest}
\usetikzlibrary{shadows.blur}
\usepackage{todonotes}
\usepackage{wrapfig}
\usepackage{bibunits}

\AtBeginDocument{%
  \providecommand\BibTeX{{%
    \normalfont B\kern-0.5em{\scshape i\kern-0.25em b}\kern-0.8em\TeX}}}

\begin{document}


\title{Domain Specialization as the Key to Make Large Language Models Disruptive: A Comprehensive Survey}

\author{Chen Ling}
\email{chen.ling@emory.edu}
\affiliation{%
  \institution{Emory University}
  \city{Atlanta}
  \state{GA}
    \country{USA}
}
\affiliation{%
  \institution{NEC Labs America}
  \city{Princeton}
  \state{NJ}
    \country{USA}
}
\author{Xujiang Zhao}
\authornote{Equal Contribution.}
\authornote{Corresponding author.}
\email{xuzhao@nec-labs.com}
\affiliation{%
  \institution{NEC Labs America}
  \city{Princeton}
  \state{NJ}
    \country{USA}
}
\author{Jiaying Lu}
\authornotemark[1]
\email{jiaying.lu@emory.edu}
\affiliation{%
  \institution{Emory University}
  \city{Atlanta}
  \state{GA}
    \country{USA}
}
\author{Chengyuan Deng}
\authornotemark[1]
\email{cd751@rutgers.edu}
\affiliation{%
  \institution{NEC Labs America}
  \city{Princeton}
  \state{NJ}
    \country{USA}
}
\affiliation{%
  \institution{Rutgers University}
  \city{New Brunswick}
  \state{NJ}
    \country{USA}
}
\author{Can Zheng}
\authornotemark[1]
\email{caz51@pitt.edu}
\affiliation{%
  \institution{NEC Labs America}
  \city{Princeton}
  \state{NJ}
    \country{USA}
}
\affiliation{%
  \institution{University of Pittsburgh}
  \city{Pittsburgh}
  \state{PA}
    \country{USA}
}
\author{Junxiang Wang}
\authornotemark[1]
\email{junwang@nec-labs.com}
\affiliation{%
  \institution{NEC Labs America}
  \city{Princeton}
  \state{NJ}
  \country{USA}
}
\author{Tanmoy Chowdhury}
\authornotemark[1]
\email{tchowdh6@gmu.edu}
\author{Yun Li}
\email{yli38@gmu.edu}
\affiliation{%
  \institution{George Mason University}
  \city{Fairfax}
  \state{VA}
  \country{USA}
}
\author{Hejie Cui}
\email{hejie.cui@emory.edu}
\affiliation{%
  \institution{Emory University}
  \city{Atlanta}
  \state{GA}
  \country{USA}
}
\author{Xuchao Zhang}
\email{xuchaozhang@microsoft.com}
\affiliation{%
  \institution{Microsoft}
  \city{Redmond}
  \state{WA}
  \country{USA}
}
\author{Tianjiao Zhao, Amit Panalkar, Dhagash Mehta, Stefano Pasquali }
\affiliation{%
  \institution{Blackrock, Inc.}
  \city{Atlanta}
  \state{GA}
  \country{USA}
}
\author{Wei Cheng, Haoyu Wang, Yanchi Liu, Zhengzhang Chen}
\affiliation{%
  \institution{NEC Labs America}
  \city{Princeton}
  \state{NJ}
  \country{USA}
}
\author{Haifeng Chen, Chris White}
\affiliation{%
  \institution{NEC Labs America}
  \city{Princeton}
  \state{NJ}
  \country{USA}
}
\author{Quanquan Gu}
\email{qgu@ucla.edu}
\affiliation{%
  \institution{University of California, Los Angeles}
  \city{Los Angeles}
  \state{CA}
  \country{USA}
}
\author{Jian Pei}
\email{j.pei@duke.edu}
\affiliation{%
  \institution{Duke University}
  \city{Durham}
  \country{USA}
  \state{NC}
}
\author{Carl Yang}
\email{j.carlyang@emory.edu}
\author{Liang Zhao}
\authornotemark[2]
\email{liang.zhao@emory.edu}
\affiliation{%
  \institution{Emory University}
  \city{Atlanta}
  \country{USA}
  \state{GA}
}

\renewcommand{\shortauthors}{Ling, et al.}

\begin{abstract}
  Large language models (LLMs) have significantly advanced the field of natural language processing (NLP), providing a highly useful, task-agnostic foundation for a wide range of applications. However, directly applying LLMs to solve sophisticated problems in specific domains meets many hurdles, caused by the heterogeneity of domain data, the sophistication of domain knowledge, the uniqueness of domain objectives, and the diversity of the constraints (e.g., various social norms, cultural conformity, religious beliefs, and ethical standards in the domain applications). Domain specification techniques are key to make large language models disruptive in many applications. Specifically, to solve these hurdles, there has been a notable increase in research and practices conducted in recent years on the domain specialization of LLMs. This emerging field of study, with its substantial potential for impact, necessitates a comprehensive and systematic review to better summarize and guide ongoing work in this area. In this article, we present a comprehensive survey on domain specification techniques for large language models, an emerging direction critical for large language model applications. First, we propose a systematic taxonomy that categorizes the LLM domain-specialization techniques based on the accessibility to LLMs and summarizes the framework for all the subcategories as well as their relations and differences to each other. Second, we present an extensive taxonomy of critical application domains that can benefit dramatically from specialized LLMs, discussing their practical significance and open challenges. Last, we offer our insights into the current research status and future trends in this area.
\end{abstract}


\maketitle

\section{Introduction}\label{sec1}

The evolution of natural language processing (NLP) and artificial intelligence (AI) models has witnessed a remarkable trajectory, beginning with the rule-based systems of the 1950s and 1960s, transitioning to statistical models in the 1990s, followed by the emergence of neural networks in the 2010s. Owing to the success of self-attention and Transformer-based neural network architecture \cite{vaswani2017attention}, Pre-trained Language Models (PLMs) emerged and swiftly gained popularity in the late 2010s due to their ability to learn universal language representations from large-scale data in an unsupervised manner, which can be beneficial for many downstream NLP tasks such as commonsense reasoning \cite{yang2023harnessing}, multiple-choice question answering \cite{robinson2022leveraging}, and story generation \cite{cao2023comprehensive}, while avoiding training new models from scratch. In the last few years, with the fast growth of large corpus and hardware capacities, researchers have found scaling up model and training data can continuously improve the model capacity, following the scaling law \cite{kaplan2020scaling}, eventually resulting in Large Language Models (LLMs) \cite{wei2022emergent}, such as GPT-3 \cite{brown2020gpt3} (175B parameters), PaLM \cite{chowdhery2022palm} (540B parameters), and LLaMA \cite{touvron2023llama} (65B parameters). LLMs, significantly outperforming smaller models in understanding and generating human-like text, have emerged as a promising AI research trend. Their potential to revolutionize natural and social sciences through efficient literature analysis, novel hypothesis generation, and complex data interpretation could accelerate research, enhance the discovery process, and facilitate interdisciplinary collaboration.

While LLMs hold great promise as general task solvers, effectively extending their functionality beyond mere ``chatbot'' roles poses significant challenges. This has led to the emergence of ``domain specialization of LLMs''.  Specifically, \textbf{domain specialization} of Large Language Models (LLMs) is defined as \emph{the process of customizing general-purpose LLMs according to specific domain contextual data, augmented by domain-specific knowledge, optimized by the domain's objective, and regulated by domain-specific constraints.} This shift towards domain specialization of LLMs is motivated by several compelling reasons. First, there are significant differences in conversation and language styles in different fields, roles, and tasks ranging from medical prescriptions to legal sentences, to online chatting, etc. The acquisition of such capabilities and experience even require human beings many years of training, a lot of which are hands-on and proprietary. Moreover, different fields, institutions, and teams have their own ``business models’’ about which response will maximize their own utility function for their tasks, which is not directly replaceable by a single general-purpose LLMs solver with no customization. More importantly, the requirement of domain knowledge for professional-level usage also need to be very in-depth, in-real-time, and accurate, none of which can be easily achieved by pre-trained LLMs. Many domain knowledge resources are proprietary assets and core competitiveness of the organizations that can never be leaked to general-purpose LLMs. Last but not the least, languages are constrained by social norms, cultural conformity, religious beliefs, legal requirements, and ethical practice, all of which are changing parameters in different locations, countries, populations, races, communities, etc., which make general-purpose LLMs impossible to be a one-fits-all solver without any customization.

Domain Specialization of LLMs is a critical yet challenging problem that requires inventing and integrating effective techniques to address the serious challenges. Particularly, there are three significant challenges.

\textbf{Challenge 1: Difficulty keeping an LLM updated with the latest knowledge.}
The power of LLMs is attributed mainly to their massive training corpus. Yet, it also indicates LLMs tend to have a knowledge cut-off and lack sufficient access to the latest information, events, or discoveries. In many specialized domains, new discoveries, regulations, and best practices continuously emerge, making it difficult for LLMs to stay up-to-date. For instance, more than 30 thousand mainstream news articles are published every day \cite{wang2022learning}. For social media analysis and fact-checking, LLMs may not handle them since the knowledge extracted from the training corpus is offline. This indicates that regular re-training or continuous learning mechanisms are required to maintain LLMs' relevance and accuracy in these dynamic fields. However, ensuring the model freshness can be resource-intensive, as it necessitates continuous high-quality and up-to-date data collection, processing, and computationally intensive model re-training. 

\textbf{Challenge 2: Difficulty in learning all specialized knowledge of different domains in one LLM.} LLMs, by default, possess general knowledge across a wide range of topics and may have seen and obtained specific knowledge for most domains. However, more popular or widely-discussed topics may be over-represented, while very domain-specific topics can usually be under-represented, which makes it difficult to be effectively learned for domain-specific tasks. In addition, domain-specific tasks often involve complex concepts, specialized terminology, and intricate relationships between entities. Without proper guidance, LLMs may generate plausible-sounding but inconsistent answers to similar queries (i.e., LLM's hallucination) or slightly rephrased questions \cite{bang2023multitask}. This issue arises because LLMs are designed to predict the most likely word sequences based on the input rather than providing a definitive answer based on a structured knowledge base. Researchers have found users can guide the model to produce more relevant, accurate, and task-specific responses, enhancing the overall utility and effectiveness of AI systems across numerous domains by providing LLMs with a few task-specific demonstrations \cite{wei2022emergent}. Nevertheless, providing LLMs with adequate demonstrations is not trivial since user instructions can often be vague, incomplete, or ambiguous, making it difficult to discern the intended meaning or desired outcome. Let alone LLMs tend to have a finite context window, typically determined by the maximum token length they can process  (e.g., ChatGPT can only handle $4097$ tokens). 

\textbf{Challenge 3: Intensive model and computational complexity required for downstream task learning.} To better adapt to specific domain applications, downstream task learning is historically a commonly used practice to specialize language models. However, different from traditional language models, adapting an LLM to downstream tasks needs vast amounts of high-quality, task-specific data. Acquiring, cleaning, and pre-processing such data can be time-consuming and resource-intensive. Moreover, the sheer complexity of LLMs makes it challenging to identify the most appropriate down-stream task learning strategy, as the choice of hyperparameters, learning rate, and training duration can significantly impact the model's performance. Chen et al. \cite{chen2020recall} have also discussed down-stream task learning for LLMs may lead to severe \textit{catastrophic forgetting} since the LLM with a complex architecture is more likely to forget previously learned knowledge and overfits to target domains. In addition to the data requirement and complex model architecture, LLMs typically consist of billions of parameters, e.g., both Generative Pre-trained Transformer 3 (GPT-3) \cite{brown2020gpt3} and Pathways Language Model (PaLM)~\cite{chowdhery2022palm} contains more than 100 billion parameters, which require substantial computational power to train. Fine-tuning or re-training these models necessitates access to high-performance GPUs or specialized hardware, such as TPUs, which can be expensive and difficult to obtain, especially for individual researchers or smaller organizations.

Over the past few years, significant research has been conducted on domain specialization techniques for LLMs. Many methods focus on generic technical contributions, adaptable to specific domains with minor modifications and access to domain-specific information. However, cross-referencing these techniques across different application domains remains a challenge, as does the absence of a systematic standardization and summary of methods for evaluating various domain specialization techniques. This lack of clarity creates obstacles for non-AI professionals and obfuscates existing bottlenecks, pitfalls, open problems, and potential future research directions. To surmount these obstacles and harness artificial intelligence for more effectively accomplishing tasks across various domains, this survey paper offers a comprehensive and systematic review of the current state-of-the-art LLM domain specialization. The major contributions of this paper include:
\begin{itemize}
    \item \textbf{A systematic categorization and taxonomy of LLMs domain specialization techniques}: We comprehensively classify existing methods based on different levels (i.e., black-box, grey-box, and white-box) of accessibility to the LLM and organize their corresponding techniques into a taxonomy. We discuss details, relationships, pros, and cons among different subcategories. The proposed taxonomy is designed to assist domain experts in identifying the most suitable techniques for their target problem settings.
    \item \textbf{A comprehensive categorization and summarization of major application domains}: We debut the taxonomy of representative application domains that domain-specialized LLMs can enhance. The practical significance and open challenges for each application domain or subdomain are elucidated, allowing for easy mapping to the proposed technique taxonomy. Researchers and various domain experts could cross-reference additional application domains for evaluating their newly proposed methods while expanding their advanced techniques to encompass new application domains.
    \item \textbf{An insightful discussion of the current status of research in this area and future trends}. Based on the comprehensive and systematic survey and investigation of existing domain specialization techniques and applications, an overall picture and trends of LLM domain specialization have been outlined and discussed. The open challenges and  The paper concludes by presenting fresh insights into the bottlenecks, open problems, as well as a discussion of possible future directions.
\end{itemize}

\subsection{Related Surveys}
This section briefly outlines previous surveys relevant to the domain specialization of LLMs in three categories: (1) fundamental overview of PLMs and LLMs; (2). techniques on domain adaptation and generalization of PLMs; and (3). Specializing language models for specific domains.

\paragraph{Fundamental overview of PLMs and LLMs.} 
While comprehensive reviews \cite{qiu2020pre, min2021recent} of PLMs and their use in diverse NLP tasks exist, they don't necessarily apply to LLMs due to differences between the two. Given the recent growth in popularity and effectiveness of LLMs, several review papers have emerged, addressing various LLM aspects. Some focus on fundamental LLM components \cite{zhao2023survey,yang2023harnessing,liu2023summary}, others on the history and potential applications of generative AI \cite{cao2023comprehensive,zhang2023complete}, and a few \cite{mialon2023augmented} on enhancing LLMs with reasoning capabilities. However, a comprehensive review and technical taxonomy of LLM domain specialization are yet to be provided.

\paragraph{Domain adaptation and generalization of PLMs.} 
Surveys \cite{guo2022domain,ding2022delta} examine how to effectively and efficiently adapt PLMs to specific domains, such as adding a layer to the model or updating the model parameters. However, most of these techniques don't apply to LLMs because of the inaccessibility of their architecture and parameter space. Also, updating knowledge in LLMs is challenging due to computational costs and the need for efficient optimization strategies.

\paragraph{Specializing language models for specific domains.} 
Recent review papers have emphasized the benefits and necessity of customizing LLMs for specific domains. Risks linked with applying generic LLMs to areas like medical education have been noted in \cite{sallam2023utility,cui2023survey}, including lack of originality and inaccuracies. Practical considerations for legal domain-specific language models have also been suggested in \cite{shaghaghian2020customizing}. In the finance sector, initial steps towards a finance-specialized LLM have shown improved performance on financial tasks without compromising general benchmarks \cite{wu2023bloomberggpt}. These advances highlight the need for a comprehensive review and technical taxonomy of domain specialization techniques to assist different sectors in effectively employing LLMs for their unique tasks.



\section{Taxonomy of Domain Specialization }\label{sec: preliminary}
Large language models are typically referred to as large-scale pre-trained language models (PLMs) based on the Transformer architecture \cite{min2021recent,qiu2020pre}. Empirical evidence suggests that scaling pre-trained language models, such as increasing the model size or data size, frequently results in enhanced model capacity for downstream tasks. In this section, we begin by reviewing the fundamental concepts of PLMs and proceed to present a comprehensive taxonomy of existing techniques aimed at specializing large language models for specific domains.
 
\subsection{Background}
 Specifically, PLM is a type of neural network pre-trained on a large corpus of text data to learn linguistic patterns, structures, and semantics. The input and output of PLMs can be described as follows. In LLMs, the \textit{input} is a text sequence that serves as context for understanding and processing. To clarify the task, a \textit{prompt}, an additional sentence or query, is often included. These prompts, designed based on the NLP task, provide a premise or task explanation. For instance, in text summarization, a prompt like "Summarize the key points in the following passage:" could precede the input passage. The \textit{output} is the text sequence or prediction generated in response to the input. Depending on the task, this could be an answer to a question or a sentiment label, and may require post-processing like token decoding or label extraction for final presentation.
As LLMs are typically scaled-up versions of PLMs, they follow the similar architecture design of PLMs, which come in three main flavors: \textit{encoder-only}, \textit{encoder-decoder}, and \textit{decoder-only} architectures. This brief introduction will provide an overview of these PLMs architectures, and discuss their differences and commonalities. 
\begin{itemize}
    \item \textit{Encoder-only Language Models} process input text into vector representations without an explicit decoding phase to generate new text. Instead, they transform and embed text into a high-dimensional space. These models are primarily designed to capture and understand the patterns and semantics in the input data. They are extensively used for tasks such as text classification, sentiment analysis, and clustering. One of the notable examples is BERT~\cite{devlin2018bert}, which extracts context-rich embeddings for downstream tasks by pre-training on a masked language modeling objective.
    \item \textit{Encoder-Decoder Language Models} consist of an encoder that processes input text into vector representations and a decoder that generates output text from these representations. They employ cross-entropy loss as the objective function, comparing the actual and predicted target sequences. These PLMs are often used for sequence-to-sequence tasks like machine translation and summarization, with T5~\cite{raffel2020exploring} being a notable example.
    \item \textit{Decoder-only Language Models}, like GPT~\cite{radford2018improving}, are autoregressive language models that generate the next word in a sequence based on previous words. They map a sequence of tokens to a vector representation and generate contextually relevant content autoregressively, calculating the probability of the next token based on the context. This autoregressive modeling approach is particularly suitable for text-generation tasks.
\end{itemize}

\subsection{Taxonomy of Domain Specialization Techniques}\label{sec: taxonomy}

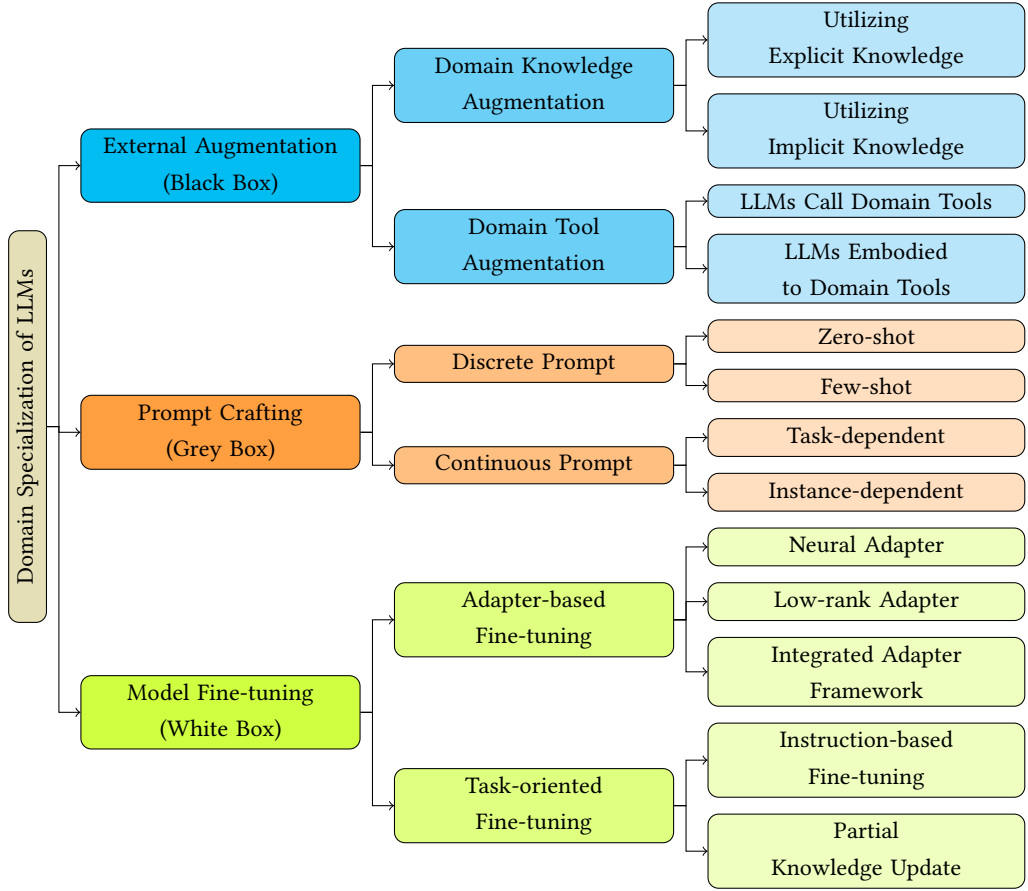
\begin{figure}[H]
\centering
\forestset{
  L1/.style={fill=white, draw=black, rotate=90, text width=5cm},
  L2/.style={fill=white, draw=black, text width=4cm},
  L3/.style={fill=white, draw=black, text width=3.5cm},
  L4/.style={fill=white, draw=black, text width=4cm},
}
\begin{forest}
for tree={draw,rounded corners,grow'=0,text width=2cm, text centered,edge+={->}},
forked edges,
    [,phantom,
    [,tier = a,draw=none, text width=0cm, for tree={color=gray, text centered, no edge, draw=none}]
      [Domain Specialization of LLMs, L1, fill=olive!25
      [External Augmentation\\(Black Box), L2, fill=cyan!75, tier=b, text width=3.5cm,
        [Domain Knowledge\\Augmentation, L3, fill=cyan!50, tier=c,
            [Utilizing \\Explicit Knowledge, fill=cyan!25,text width=4cm,tier=d]
            [Utilizing \\Implicit Knowledge, fill=cyan!25,text width=4cm]
        ]
        [Domain Tool\\Augmentation, L3, fill=cyan!50, tier=c,
            [LLMs Call Domain Tools,fill=cyan!25,text width=4cm]
            [LLMs Embodied to Domain Tools,fill=cyan!25,text width=4cm]
        ]
      ]
      [Prompt Crafting\\(Grey Box), L2, fill=orange!75, tier=b, text width=3.5cm,
        [Discrete Prompt, L3, fill=orange!50, tier=c,
            [Zero-shot, L4, fill=orange!25]
            [Few-shot, L4,  fill=orange!25]
        ]
        [Continuous Prompt, L3, fill=orange!50, tier=c,
            [Task-dependent, L4, fill=orange!25]
            [Instance-dependent, L4,  fill=orange!25]
        ]
      ]
      [Model Fine-tuning\\(White Box), L2, fill=lime!75, tier=b, text width=3.5cm,
        [Adapter-based\\Fine-tuning, L3, fill=lime!50, tier=c,
            [Neural Adapter, L4, fill=lime!25]
            [Low-rank Adapter, L4, fill=lime!25]
            [Integrated Adapter Framework, L4, fill=lime!25]
        ]
        [Task-oriented\\Fine-tuning, L3, fill=lime!50, tier=c, 
            [Instruction-based\\ Fine-tuning , L4, fill=lime!25]
            [Partial\\ Knowledge Update, L4, fill=lime!25]
        ]
      ]
      ]
    ]
\end{forest}
\normalsize
\caption{The taxonomy for current techniques on LLM domain specialization.}
\vspace{-3mm}
\label{fig:taxonomy}
\end{figure}

\begin{figure*}[!t]
    \centering
    \includegraphics[width=\textwidth]{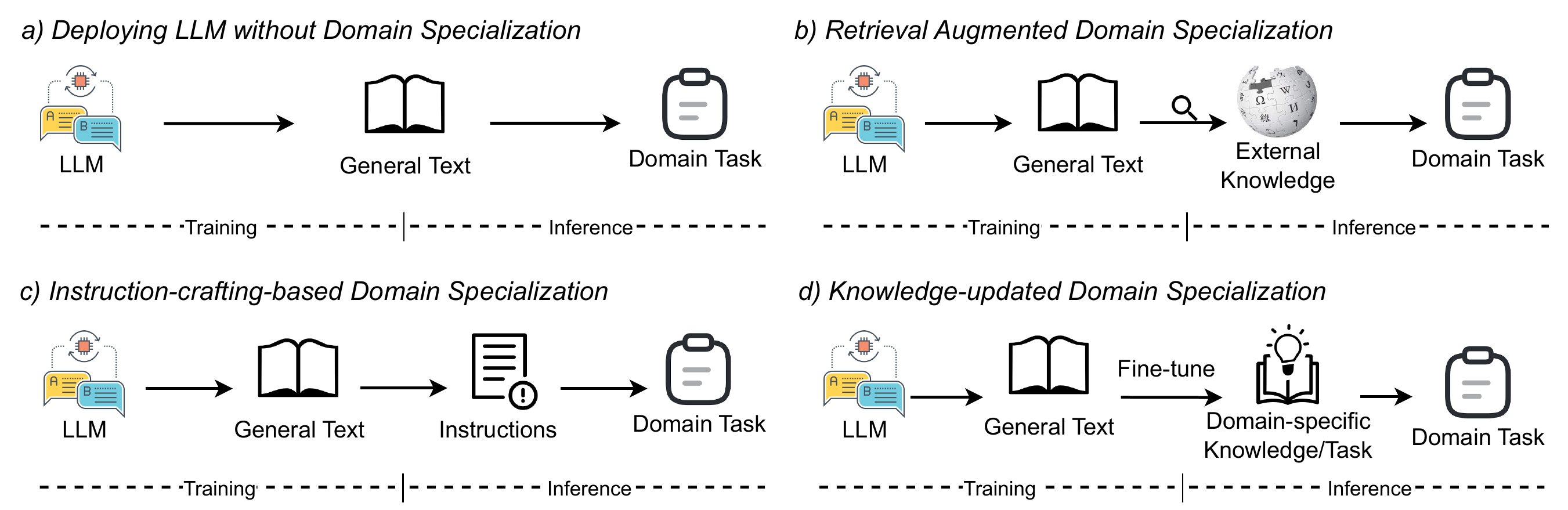}
    \vspace{-5mm}
    \caption{This exposition discusses different approaches for tailoring LLMs to domain-specific tasks: (a) using an LLM trained on general corpora without modifications, (b) enhancing the LLM's performance through retrieving relevant external knowledge, (c) utilizing domain-specific and task-relevant instructions to improve LLM's capabilities, and (d) updating the LLM's internal knowledge with domain-specific text and tasks.}
    \vspace{-5mm}
    \label{fig: example}
    \end{figure*}

Domain specialization of LLMs can be understood as tailoring broad, universally-trained LLMs to operate optimally within a specific field or domain. To tackle the three challenges of domain specialization mentioned in Section~\ref{sec1}, respectively, the approaches in LLM domain specialization can be categorized into three corresponding classes of approaches: \emph{external augmentation}, \emph{prompt crafting}, and \emph{model fine-tuning}. These classes correspond to \textit{assumptions} of different levels of accessibility to LLMs, namely, no access (black box), partial access (grey box), and full access (white box). The black box \textit{assumption} typically indicates we only have access to the model API (e.g., ChatGPT) without knowing any information but the generated output; the grey box \textit{assumption} denotes we have limited information (e.g., the probability of generated tokens in GPT-3 API), such information can guide us to design and fine-tune a suitable prompt to elicit domain knowledge better; and the white box \textit{assumption} indicates we have full access to the LLM (e.g., LLaMA and its variants), including the parameter setting, training data, and the model architecture. 

Other than the LLM accessibility-based taxonomy, one way to categorize LLM domain specialization methods is based on the training strategy used, such as fine-tuning an existing model with domain-specific data, training a model from scratch specifically for the domain, or employing a mixed training strategy. An additional taxonomy could be based on the intervention level: pre-training intervention involves modifying the pre-training process to encourage domain-specific knowledge, the fine-tuning intervention involves adaptations during the fine-tuning stage, and inference-time intervention involves modifying the model's behavior during the actual application to generate more domain-specific outputs. Furthermore, the taxonomy can be established based on the evaluation and feedback mechanism: fixed evaluation sets a constant benchmark, dynamic evaluation involves continuous performance assessment with changing benchmarks, and user feedback-based evaluation uses direct user input as a signal to specialize the model's responses.

In this survey, we categorize existing approaches based on the LLM's \textbf{accessibility} and provide an overview of each approach in Figure \ref{fig: example}. To be more specific, \textit{1) External augmentation (black box)} does not necessarily require access to the LLM's inner parameter space, making it the most accessible for users with limited resources (e.g., computational resources, domain-specific data). As shown in Figure \ref{fig: example} (b), by using external resources or tools, domain-specific knowledge is incorporated into the input prompt, generated output, or both, effectively adapting the LLM's performance without modifying its internal structure. \textit{2) prompt crafting (grey box)} involves designing various types of prompts by accessing the gradient or loss values of LLMs, allowing for finer control over the model's behavior. \textit{3) model fine-tuning (white box)} demands the most access and resources, as it involves updating the LLM's parameters to incorporate domain-specific knowledge directly into the model. (Figure \ref{fig: example} (d)).

\paragraph{Relations between approaches in different categories}
\begin{itemize}
    \item \textbf{Different levels of specialization}: Each approach operates at a different level of specialization (i.e., black box, grey box, and white box). Augmenting with external knowledge provides a focused injection of domain-specific information while prompt engineering works at the input level, shaping the model's inference process. Fine-tuning modifies the LLM's internal parameters, leading to more profound changes in the model's behavior.
    \item \textbf{Trade-offs}: The approaches exhibit different trade-offs regarding computational cost, ease of implementation, and generalization. Augmenting with external information and crafting task-specific instructions are often less computationally expensive than knowledge updates of LLMs but may not yield the same level of performance improvement. Fine-tuning and neural adapters can provide more substantial performance gains but can be more challenging to implement and may suffer from reduced generalization capabilities if overfitting occurs.
    \item \textbf{Complementary nature}: The three approaches can be used independently or in combination to achieve better performance on domain-specific tasks. For instance, external knowledge can be integrated with a fine-tuned LLM to leverage both specialized knowledge and optimized parameters. Similarly, carefully designed prompts can be used alongside neural adapters to guide the model's output while taking advantage of the newly learned domain-specific knowledge.
\end{itemize}

\paragraph{Common Framework.}
 Researchers can utilize these methods independently or in combination to achieve optimal performance on specific tasks while considering the unique requirements and constraints of each approach.  In this paper, we provide a common framework underlying the black box, grey box, and white box methods for domain specialization of LLMs, which is a process consisting of four core stages: \textit{Definition}, \textit{Augmentation}, \textit{Optimization}, and \textit{Evaluation}.
 \begin{enumerate}
     \item \textbf{Definition}: This is the first step where the specific domain, the objectives within that domain, and any constraints are clearly defined. Whether we fine-tune a model (white box), craft prompts (grey box), or augment inputs/outputs (black box), it requires a clear understanding of the domain we are specializing for. This also helps in identifying the specific data, knowledge, and resources relevant to the domain that could be used in the following steps.
     \item \textbf{Augmentation}: This stage involves the incorporation of domain-specific knowledge into the model, or its inputs/outputs. In a white box approach, this could involve fine-tuning the model with domain-specific data. For a grey box approach, it might involve using gradients or loss values to craft prompts that steer the model toward domain-specific responses. In a black box method, it could involve using external tools or resources to modify the input prompt or the generated output to make it more domain-specific.
     \item \textbf{Optimization}: Once the model or its inputs/outputs are augmented with domain knowledge, the next step is to optimize the model's performance to best fulfill the domain objectives. This can be done through methods like gradient descent for a white box approach, prompt engineering for a grey box approach, or post-processing and filtering of outputs for a black box approach.
     \item \textbf{Evaluation}: The final stage involves testing the specialized model's performance against predefined benchmarks, gathering feedback, and refining the model based on this feedback. This could involve running the model on a domain-specific test set or getting feedback from domain experts.
 \end{enumerate}

\section{External Augmentation for Domain Specialization}\label{sec: augmentation}

Retrieval augmentation aims to enhance LLMs by retrieving relevant information from external sources, without fine-tuning model parameters. There are two primary categories: (1) \textbf{Domain Knowledge Augmentation}, where LLMs are provided with domain-specific context from an external knowledge source, and (2) \textbf{Domain Tool Augmentation}, which integrates LLMs with external systems or tools, often via APIs. Domain Knowledge Augmentation supplements the model's responses with external information, while Domain Tool Augmentation expands the model's capabilities for tasks it couldn't perform otherwise. Domain Knowledge improves depth and accuracy within a specific field, while Domain Tools allow the model to perform tasks beyond its inherent abilities. This section discusses both approaches, their limitations, and their advantages.



\subsection{Domain Knowledge Augmentation}
Domain knowledge, in the broadest sense, is a comprehensive understanding of a specific field or subject area. It includes concepts, principles, facts, and patterns that are unique to a particular domain. The knowledge can be represented in various forms, including a set of documents, a domain-specific knowledge graph, or a neural network that contains parametric domain knowledge. Domain knowledge augmentation in LLM specification refers to the process of enriching an LLM's performance in specific domains by incorporating additional information from domain knowledge. Two categories of external knowledge typically can facilitate LLMs in their domain specialization: \textbf{explicit knowledge} refers to knowledge that is clearly defined, easily expressed, and structured in a manner that can be directly understood and utilized; and \textbf{implicit knowledge} refers to knowledge that is not directly stated or easily expressed, but is embedded within the data or the system, often in a latent, non-obvious form.


\subsubsection{Utilizing Explicit Knowledge with LLM} \label{ssec:explict_knowledge_LLM}
A conventional method for customizing language models to domain-specific tasks is to retrieve domain-specific information from external context. When presented with an explicit knowledge source containing domain-specific information, it is crucial for LLMs to prioritize the context if the data source holds task-relevant details that contradict the model's memorized knowledge. This strategy ensures that model predictions are anchored in the context, allowing for the refinement or correction of specific model predictions without the need for frequent retraining.

Current techniques often employ a neural retriever to acquire task-relevant information from either a large corpus (e.g., Wikipedia) or a knowledge base (e.g., Wikidata) ~\cite{lewis2020retrieval,singh2021end,liu2022relational,dai2022promptagator,izacard2022few,he2022rethinking,lu23HiPrompt,li2022large,borgeaud2022improving}. Specifically, given a task-specific query, early works \cite{lewis2020retrieval,singh2021end,liu2022relational,borgeaud2022improving} designed neural retrievers to vectorize the query and all information in the external knowledge source to search for relevant information based on various similarity metrics (e.g., cosine similarity) in the latent space. The searched information can then be concatenated with the query for downstream tasks. With the prevalence of LLMs, researchers have been using LLMs to replace the neural network-based retriever \cite{dai2022promptagator,izacard2022few,ram2023context}, and one work \cite{izacard2022few} demonstrated that coupling a rather lightweight LLM (around 11 billion parameter size) with an external knowledge base can achieve similar performance when using a 540B LLM (i.e., PaLM). Furthermore, in order to enhance the transparency and explainability of the retrieval, He et al. \cite{he2022rethinking} proposed to leverage LLMs to decompose the information retrieval process with detailed reasoning steps, and Lu et al. \cite{lu23HiPrompt} explored to utilize LLMs to verify whether information obtained by a pre-trained neural-network-based retriever is relevant or not.

\subsubsection{Utilizing Implicit Knowledge with LLM}
Implicit domain knowledge in machine learning refers to latent, non-obvious information embedded within data or the system, often represented as vectorized knowledge or embeddings learned during pre-training. Such embeddings capture intricate data patterns, symbolizing domain knowledge in an abstract form. Previous research \cite{grave2016improving,merity2016pointer,feng-etal-2017-memory,wan2022g} suggests the use of attention mechanisms to enable PLMs to retrieve task-related information from this implicit knowledge. These studies transform task-specific queries into latent embeddings, calculating attention scores between the query vector and each knowledge entry. A softmax function is used to generate a weight or probability distribution across all knowledge entries concerning the input query. The retrieved memory vector is then obtained via a weighted sum of the memory entries, using attention weights. This method enhances traditional neural networks with implicit knowledge, permitting the model to access relevant, current information during inference.

While LLMs can store a substantial amount of information in their parameters to generate high-quality responses, augmentation with implicit knowledge isn't always required. Unlike explicit knowledge, implicit knowledge requires extra processing, such as transforming domain-specific data into latent vectors, making it less practical. Despite the limited work in augmenting LLMs with implicit knowledge, researchers are exploring its potential, including its use in storing instructional knowledge about a domain. This approach involves creating an instruction cycle that retrieves the next input prompt from implicit knowledge, parses the LLM's output to recover variable assignments, and stores these back into the memory for retrieving the next instruction. Augmenting LLMs with this instruction cycle allows them to process large inputs and potentially solve complex domain-specific problems \cite{schuurmans2023memory}.

\subsubsection{Open Challenges} 
By incorporating external knowledge, LLMs function like librarians, finding relevant information without needing to memorize all resources. This enhances performance in specialized tasks without extensive retraining, enabling more adaptable and efficient AI systems capable of lifelong learning and knowledge updating. However, augmenting LLMs with external knowledge for domain-specific tasks presents several open challenges.
\begin{enumerate}
    \item \textit{Seamless integration}: Seamless integration of external knowledge into LLMs is crucial, whether the knowledge is explicit or implicit. Existing methods typically concatenate retrieved knowledge to the LLM's input or intermediate layers. However, it's important for the LLM to have the option of accepting or rejecting retrieved information, given that such information may be incomplete or conflicting.
    \item \textit{Scalability and adaptability}: Designing systems capable of scaling to manage large amounts of domain-specific data and adapting to new or changing information is challenging. With rapidly expanding knowledge bases, computing pairwise knowledge similarity will become increasingly computationally unfeasible.
\end{enumerate}


\subsection{Domain Tool Augmentation}
Domain tools refer to specialized software, libraries, or frameworks that are developed specifically for a particular domain or field (\textit{e.g.}, NCBI Web APIs for genomics question answering~\cite{jin2023genegpt}, automated formal theorem prover for mathematical proofs~\cite{jiang2023draft}, sandbox environment for social behavior simulation~\cite{park2023generative}, etc.). These tools are designed to handle domain-specific tasks, data, or knowledge effectively, which often incorporate algorithms, techniques, or data structures that are tailored to the unique requirements of that domain. However, the utilization of these domain tools often demands strict adherence to input formats or extensive training, making them less accessible to general users. On the other hand, LLMs are artificial general intelligence models that exhibit intelligence and cognitive capabilities across a wide range of tasks and domains. Despite their versatility, current LLMs are constrained in tasks that require domain specialization. These limitations~\cite{mialon2023augmented,schick2023toolformer} include (1) unstable result formats depending on the random seeds, generation hyperparameters, and input contents~\cite{schick2020exploiting}; (2) inability to access up-to-date information~\cite{nakano2021webgpt} since LLMs are solely capable of acquiring information from their training data; (3) a tendency to make up facts observed by researchers~\cite{ji2023hallucination}; (4) lack of precision in certain tasks such as arithmetic~\cite{gao2022pal}.

Researchers propose a collaborative integration approach to overcome the limitations of solely using either domain tools or LLMs for complex domain-specific tasks. This approach combines the strengths of both, utilizing domain-specific knowledge, algorithms, and functionalities from the tools, while offering a user-friendly interface through LLMs. This collaboration optimizes the use of domain-specific resources and eases user engagement by allowing LLMs to guide effective tool usage.
\paragraph{LLMs Call Domain Tools}

\begin{figure*}[ht!]
    \centering
    \includegraphics[width=\textwidth]{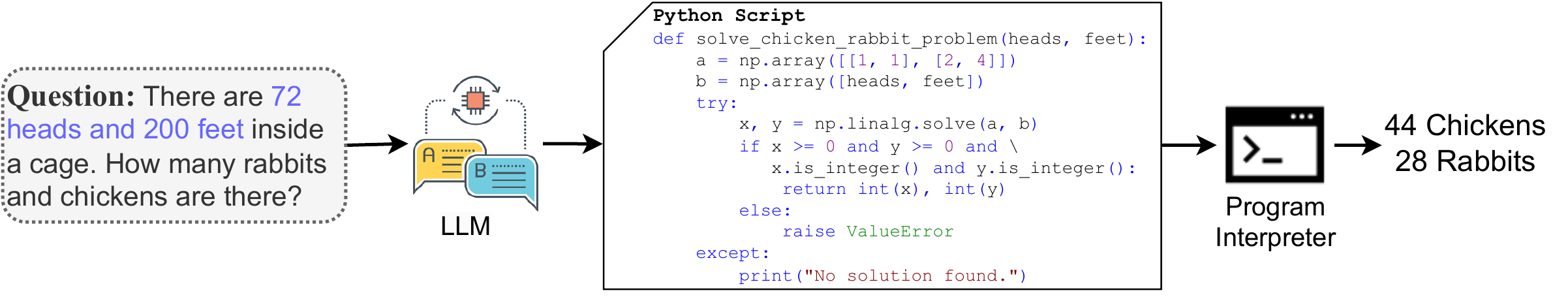}
    \vspace{-5mm}
    \caption{A toy example for LLMs call domain tools.}
    \vspace{-3mm}
    \label{fig:LLM_call_tool}
\end{figure*}

One straightforward way for domain tool augmentation is to allow LLMs to call domain tools. Essentially, this type of approach follows a multi-stage pipeline, given an LLM $f_{\Theta}(\cdot)$ and a domain tool $\mathcal{T}(\cdot)$: (1) elicit an executable command $c$ for the domain tool from the LLM by curated or constructed prompts $p$, denoted as $c = f_{\Theta}(p)$. (2) execute the command $c$ in the domain tool and get the outputs, denoted as $r=\mathcal{T}(c)$. (3) post-process the domain tool outputs by pre-defined rules or the LLM, denoted by $y=post-process(r)$.

This pipeline provides a general diagram and can be easily expanded to multi-LLMs multi-tools collaboration scenarios.
The key technical challenge is to ensure the instruction-following and validity of generated commands $c$ from LLMs, so that domain tools can accurately solve desired tasks. Most existing works propose to utilize zero-show or few-shot prompting for executable commands generation (please refer to Sec.~\ref{sec: instruction} for more details). Figure~\ref{fig:LLM_call_tool} shows a toy example, where the task is to solve an arithmetic question ``\emph{There are 72 heads and 200 feet inside a cage. How many rabbits and chickens are there?}''. To elicit LLMs to generate an executable Python program, we can formulate the prompt as ``\emph{Please write a Python script to solve the arithmetic question. Question: \{question\_text\}}''. Then, a snippet of scripts is returned by LLM as the executable command $c$ for the Python interpreter. Finally, the Python interpreter responds with the program outputs ``\emph{44, 28}'', and further post-processed into desired results ``\emph{44 Chickens and 28 Rabbits}''.

Depending on the types of domain tools, LLMs can generate corresponding commands that adhere to the syntax and format requirements to call them. 
Many domain tools provide APIs for easy and precise access. Early exploration in this direction is to elicit elicit search engine queries (\textit{e.g.}, WebGPT~\cite{nakano2021webgpt}, WizInternet~\cite{komeili2022internet}, GopherCite~\cite{menick2022GopherCite}) written in natural language or database queries (\textit{e.g.}, Binder-SQL~\cite{cheng2023binding}, DIN-SQL~\cite{pourreza2023DINSql}, BIRD~\cite{li2023BIRD}) in the programming language. Later, researchers study how to elicit LLMs to write executable codes that can be executed in program interpreters such as Python~\cite{gao2022pal,chen2022program,suris2023vipergpt}, Wolfram~\cite{chatGPTWolfram}, and so on. Other than the widely-used search queries, database queries, and Python programs, there exist many domain-specialized APIs that have unique syntax. For instance, chatGPT plugin system~\cite{chatGPTPlguin} introduces how to utilize tools for travel booking, restaurant reservation, e-commerce shopping, and workflow automation. These API calling scripts are typically generated by zero-shot or few-shot prompting techniques, as stated in the toy example.


Some complex tasks may involve more than one type of tool to accomplish. Following this vibe, researchers start to generalize LLMs as \textbf{task planners} (also mentioned as ``API selectors'' or ``controllers'') that call multiple types of domain tools. Other than generating executable commands for each used tool, these approaches focus on how to decompose a complex task into a set of concrete subtasks and how to coordinate between multiple tools. For instance, DSP~\cite{jiang2023draft} proposes a Draft, Sketch, and Prove framework for automated theorem proofs where (1) an LLM or oracle is used to draft informal proofs described in a mixture of natural and mathematical languages from input statements, (2) another LLM is used to generate formal sketch from previous informal proof, and (3) off-the-shelf prover is used to prove the open conjectures inside each formal sketch. TaskMatrix.AI~\cite{liang2023taskmatrix} proposes using LLMs to derive high-level solution outlines for domain-specific tasks, and automatically match some of the sub-tasks in the outlines to the off-the-shelf domain models/systems to complete them. HuggingGPT~\cite{shen2023hugginggpt} proposes leveraging LLMs to act as the controllers to manage existing domain models to solve complicated domain tasks. Qin et al.~\cite{qin2023tool} propose a general tool-augmented LLMs framework to decompose complex tasks into several subtasks, dynamically adjust the execution plan, and effectively finish each subtask with appropriate tools. 

\paragraph{LLMs Embodied to Domain Tools}  
LLMs can also be called by domain tools to serve as smart agents in interactive environments, namely \emph{LLMs embodied to domain tools}. 
LLMs, when embodied in interactive robots, can serve as the decision-making module for domain-specific applications. For example, \textsc{ProgPrompt}~\cite{singh2022progprompt} investigates LLMs' ability to assist robots in completing tasks, when the robot's perception module observes surrounding objects and the LLM is prompted with available action specifications. Results indicate the LLM can generate situated actions for simulated household and real-world tabletop tasks. Furthermore, Murali et al.~\cite{murali2023improving} employ LLMs as the primary component for identifying different speakers in multiparty conversations involving a social robot. The robotics community is progressively exploring these areas, studying LLM utility in human-robot interfaces, planning, grounding, and more ~\cite{wang2023describe,dasguptacollaborating,liang2022code}. 
Furthermore, researchers start to investigate how multiple LLMs can \emph{interact with the environment} or \emph{communicate and collaborate} together for real-world task-solving. Mind's eye~\cite{liu2022mind} studies how LLMs can benefit from the interaction with simulated physics engines to inject grounded rationale for physics alignment tasks. CAMEL~\cite{li2023camel} proposes a communicative agent framework to assign different roles to LLM agents so that multiple AI agents can collaboratively communicate by chatting with each other in an instruction-following fashion to solve the specified task. A recent work \cite{park2023generative} utilizes twenty-five LLMs as generative agents in a game-based sandbox environment to create believable simulations of human behavior for interactive applications. 

\subsubsection{Open Challenges}
By leveraging the power of LLMs, domain tools can assist in a variety of tasks across multiple fields, including robotics, virtual agents, and problem-solving in real-world scenarios. This allows for more intuitive and seamless human-machine collaboration, leading to increased efficiency and adaptability in tackling complex problems.

Augmenting LLMs with domain tools poses several open challenges:

\begin{enumerate}
    \item \emph{Automated integration}: At present, augmenting LLMs with domain-specific tools requires a significant amount of effort to ensure proper integration. A promising future direction involves utilizing LLMs as a unified interface through standardized protocols to connect various applications and services, thereby enabling seamless communication and interaction between them.
    \item \emph{Getting rid of domain tools}: Another direction for the future development of LLMs is to focus on creating a powerful artificial general intelligence (AGI) model that is not dependent on external tools or domain-specific knowledge. An AGI model would have the potential to revolutionize the way we use language models, enabling more complex and sophisticated tasks to be performed with greater ease and efficiency.
\end{enumerate}

\section{Prompt Crafting for Domain Specialization}\label{sec: instruction}

While LLMs trained on large-scale corpora are powerful, further pre-training on prompts can enhance their ability to adhere to user intentions and generate accurate and less toxic responses~\cite{ouyang2022InstructGPT,wang2022selfInstruct}. Prompts, or task-specific input texts designed to elicit specific model responses, help guide the LLM's content generation process and set expectations for the desired output. Approaches generally fall into two categories: (1) \textbf{Discrete Prompt} involves creating task-specific natural language instructions to prompt LLMs, eliciting domain-specific knowledge from their parameter space, and (2) \textbf{Continuous Prompt} uses learnable vectors to prompt LLMs, eliminating the need for manually designed text instructions. This section delves into both approaches and the merits and limitations of domain specialization.


\subsection{Discrete Prompt}
Recent works~\cite{brown2020gpt3,openai2023gpt4} allow LLMs to quickly adapt to unseen domains by discrete prompting, and GPT-3~\cite{brown2020gpt3} is the first work that introduces how to perform an unseen task using an LLM with zero-shot/few-shot discrete prompts without updating the LLM's inner parameter. We give a formal definition of the discrete prompt framework below.

\noindent \textbf{Problem Setup.} Given an LLM $f_{\Theta}(\cdot)$ where $\Theta$ denotes pre-trained model parameters, the task is to elicit desired output $\boldsymbol{y}$ from LLM with a discrete prompt $p$ and a test query, denoted as $\boldsymbol{y}=f_\Theta([p; \boldsymbol{c}])$, when freezing $\Theta$. It is worth noting that both $\boldsymbol{y}$, $p$ and $\boldsymbol{c}$ are sequences of tokens (i.e., natural language sentences). The rationale behind using discrete prompts is that they can serve as instructions to elicit the generalized reasoning abilities of LLMs. By following such instructions, LLMs can perform domain-specific tasks that they have not been specifically trained for. This approach allows LLMs to demonstrate their ability to apply previously learned knowledge to new and diverse situations, thus enhancing their overall effectiveness and utility.

Depending on the prompting types, discrete prompts can be divided into two categories: (1) zero-shot~\cite{ben2022pada,kojima2022zeroCoT,sanh2022multitask}, where the prompt $p$ consist of only the task description; and (2) few-shot~\cite{wei2022CoT,min2022rethinking,madaan2022CoCoGen}, where the prompt $p$ consists of the task description and few illustrative examples. The key difference between zero-shot and few-shot prompts is whether or not illustrative examples are provided.

\subsubsection{Zero-shot Discrete Prompts}
The zero-shot setting represents the cold-start scenario, where not a single supportive labeled example is available. 

\begin{wrapfigure}{r}{0.5\textwidth}
  \centering
  \vspace{-5mm}
  \begin{tcolorbox}[colback=white,colframe=blue]
  \textcolor{lightgray}{\# Task description}\newline
  Please determine if the two sentences entail, contradict, or are neutral to each other. \newline
  \textcolor{lightgray}{\# Test query}\newline
  \textbf{Premise}: She emerged vigorous with Apgar of 7 and 8.\newline
  \textbf{Hypothesis}: She had low APGAR scores.\newline
  \textbf{Answer}:\newline
  \textcolor{lightgray}{\# LLM response}\newline
  \textcolor{blue}{LLM$:$} \colorbox{cyan}{Contradiction}
  \end{tcolorbox}
  \vspace{-3mm}
  \caption{An example (adapted from~\cite{lehman2023CLM}) of zero-shot discrete prompts, where task description, and/or a test query are provided to LLMs. No illustrative examples are provided in zero-shot prompts.}
  \label{fig:zero_shot_prompt}
  \vspace{-3mm}
\end{wrapfigure}
Figure~\ref{fig:zero_shot_prompt} presents a toy example of how zero-shot discrete prompts work. The task description that compromises the prompt $p$ can be curated by human users or automatically generated by templates, where the intent of the task and the expected outcomes are described in natural language. However, as stated in ~\cite{schick2020exploiting}, post-process is sometimes required to extract the rigorous prediction results from the unbounded raw outputs. Researchers demonstrate that instruction alignment pre-training enables decent zero-shot performance on various unseen tasks~\cite{brown2020gpt3,sanh2022multitask,wei2022finetuned}, where different tasks can be represented in a unified sequence generation format.
PADA~\cite{ben2022pada} is one of the pioneering works that explore how to elicit the domain adaptation ability of LLMs for domains unseen during the training phase. PADA first generates the target domain name followed by a set of domain-related features related to the test query, and then uses them together as the prompt to predict task labels. Follow-up works explore how to utilize zero-shot discrete prompt for domain adaptation in sentiment analysis~\cite{jia2022prompt}, image classification~\cite{ge2022DAPL}, semantic segmentation~\cite{fahes2022poda}, and rumor detection~\cite{lin2022zero}.
Later, Kojima et al. \cite{kojima2022zeroCoT} extend the few-shot-Chain-of-Thoughts(Few-shot-CoT)~\cite{wei2022CoT} into zero-shot-CoT to elicit multi-step reasoning ability of LLMs. The core idea of Zero-shot-CoT is a two-stage prompting, where the 1st stage simply adds the same prompt ``\textit{Let's think step by step}'' before each answer to derive the reasoning process sentences, and the 2nd stage takes the generated reasoning sentences to generate the final answer. Zero-shot-CoT has achieved significantly stronger performance than the standard zero-shot prompting method on arithmetic, symbolic reasoning, and other logical reasoning tasks.


\subsubsection{Few-shot Discrete Prompts}

The few-shot setting reflects the characteristics of sparse training samples of many domain-specific applications (i.e., only a few annotated examples are available).

\begin{wrapfigure}{l}{0.49\textwidth}
  \centering
  \vspace{-4mm}
  \begin{tcolorbox}[colback=white,colframe=blue]
  \textcolor{lightgray}{\# Task description}\newline
  Please determine if the two sentences entail, contradict, or are neutral to each other. Below are several examples.\newline
  \textcolor{lightgray}{\# Example 1}\newline
  \textbf{Premise}: ALT, AST, and lactate were elevated as noted above.\newline
  \textbf{Hypothesis}: The patient has abnormal lfts.\newline
  \textbf{Answer}: Entailment \newline
  \textcolor{lightgray}{\# Example 2}\newline
  \textbf{Premise}: Chest x-ray showed mild congestive heart failure.\newline
  \textbf{Hypothesis}: The patient complains of cough.\newline
  \textbf{Answer}: Neutral \newline
  \newline
  \textcolor{lightgray}{\# Test query}\newline
  \textbf{Premise}: She emerged vigorous with Apgar of 7 and 8.\newline
  \textbf{Hypothesis}: She had low APGAR scores.\newline
  \textbf{Answer}:\newline
  \textcolor{lightgray}{\# LLM response}\newline
  \textcolor{blue}{LLM$:$} \colorbox{cyan}{Contradiction}
  \end{tcolorbox}
    \vspace{-3mm}
  \caption{An example (adapted from~\cite{lehman2023CLM})  of few-shot discrete prompts, where task description, illustrative examples, and/or a test query are provided to LLMs. Compared to zero-shot prompts, a few supportive labeled samples are provided in few-shot prompts. }
  \label{fig:few_shot_prompt}
    \vspace{-3mm}
\end{wrapfigure}

Figure~\ref{fig:few_shot_prompt} presents a toy example of how few-shot discrete prompts work. Different from zero-shot prompts, a few examples that further convey the task intention and provide illustrations of the desired output formats are included in the prompt $p$. Researchers have observed that few-shot prompts yield more stable output formats and more decent performance on downstream tasks~\cite{brown2020gpt3,openai2023gpt4}.
Chain-of-Thought (CoT)~\cite{wei2022CoT} improves the domain specialization ability of LLMs by introducing a series of intermediate reasoning steps for complex reasoning tasks, but it also brings extra cost in manually designing CoTs for each test example. As a follow-up, Auto-CoT~\cite{zhang2022autoCoT} eliminates manual designs by appending the ``\textit{Let's think step by step}'' prompt to the given task context and letting LLMs generate reasoning chains directly. Other than the natural language format instructions, CoCoGen~\cite{madaan2022CoCoGen} studies the programming language format instructions to tackle structured reasoning tasks.

More advanced techniques are then proposed to further improve discrete instruction of LLMs for domain specialization and customization. For instance, \emph{ensemble-based instruction}~\cite{wang2023selfconsistency,wang2022rationale,lu2022fantastically} utilizes multiple different instructions to derive multiple model outputs and then aggregates these outputs to achieve better task performance.
Another line of research proposes \emph{recursive instruction}~\cite{zhou2023leasttomost,arora2023ask,dua2022successive,yang2022seqzero,drozdov2023compositional,khot2022decomposed} to breaks down a complex unseen task into a series of subtasks that are relatively easier to solve and then employs LLMs with specific instructions for each subtask.

\subsubsection{Open Challenges}
Utilizing discrete prompts helps LLMs leverage their inherent knowledge to adapt to new and diverse situations. This approach not only demonstrates the flexibility and adaptability of LLMs but also enhances their overall effectiveness and utility across a wide range of domains and tasks. However, crafting discrete prompts of LLMs for domain specialization poses several open challenges:

\begin{enumerate}
    \item \emph{Effectiveness}: Often the discrete instructions are curated by domain experts or follow some types of templates. It is arguable whether the instructions used are the most effective ones. Therefore, there is a need for evaluation of these instructions. This can be achieved through collaboration between domain experts and data scientists, who can analyze the performance of the LLMs and adjust the instructions accordingly. An automatic evaluation would be even better.
    \item \emph{Scalability and adaptability}: Automated ways to generate and select/combine discrete instructions without excessive human intervention is another promising direction to improve discrete instructions of LLMs.
\end{enumerate}

\subsection{Continuous Prompt}
Similar to discrete prompts, the continuous prompt is a sequence of tokens proposed to attach with the input sentence and guide LLMs with extra knowledge but can be learned from the downstream dataset by continuous \emph{prompt tuning}. In this case, the continuous prompt serves as a \textit{soft} parameterized prompt instead of the hard-coded instruction as discrete language phrases. Prompt tuning is to optimize the prompt that adapts an LLM to customized tasks or domains with the preservation of LLM's general language understanding ability. Here, one can solely update prompt-related parameters, whose quantity is  around only $0.01\%$ of the total number of the LLM's parameters, while freezing the LLM itself during the fine-tuning phase. 

\begin{figure*}[ht!]
    \centering
    \includegraphics[width=0.8\textwidth]{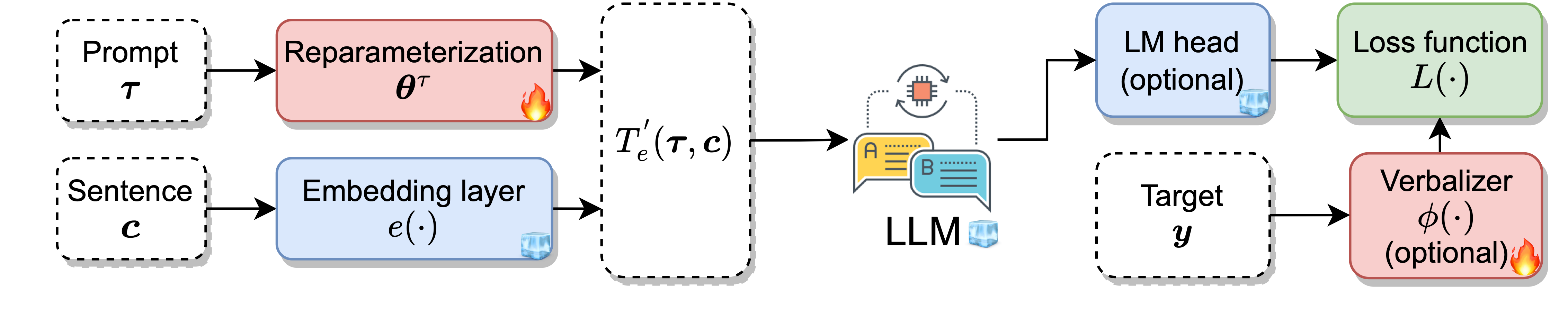}
    \vspace{-4mm}
    \caption{An illustration of soft prompt tuning. Fire icon represents tunable modules and ice icon represents that parameters of those modules are frozen during tuning. Verbalizer are only used for classification task where a mapping from class label to label words is required, which can be one-one mapping, trainable tokens \cite{hambardzumyan2021warp}, or enhanced with extra knowledge \cite{hu-etal-2022-knowledgeable}.}
    \label{fig:soft_promt}
    \vspace{-5mm}
\end{figure*}

A general framework of continuous prompt tuning (Figure \ref{fig:soft_promt}) can be concisely described in the following stages: (1) Given an input sentence $\boldsymbol{c}$ and its corresponding target $\boldsymbol{y}$, a template function $T(\cdot)$ organizes them along with a prompt $\boldsymbol{\tau}$ of length $m$ into a new sentence $T(\boldsymbol{\tau}, \boldsymbol{c})=\{\boldsymbol{\tau}_{0:i}, \boldsymbol{c}, \boldsymbol{\tau}_{i+1:m}\}$. (2) Subsequently, the sequence $T(\boldsymbol{\tau}, \boldsymbol{c})$ is mapped into an embedding space using model's input layer $\boldsymbol{e}(\cdot)$, resulting in the sequence of token embeddings: $T_e(\boldsymbol{\tau}, \boldsymbol{c})=\{\boldsymbol{e}(\tau_1),..., \boldsymbol{e}(\tau_i), \boldsymbol{e}(\omega_1),..., \boldsymbol{e}(\omega_n), \boldsymbol{e}(\tau_{i+1}),..., \boldsymbol{e}(\tau_m)\},$
where $\tau_i$ is the $i$-th token in the prompt and $T_e(\cdot)$ denotes the sequence in the embedding space. To perform prompt tuning, $\boldsymbol{\tau}$ is considered as pseudo tokens without explicit semantic meanings, and thus $\boldsymbol{e}(\tau_i)$ is replaced with a trainable tensor $\boldsymbol{h}(\tau_i)$ reparameterized by ${\boldsymbol{\theta}}^{\tau}$. This modifies the template to: $T_e^{'}(\boldsymbol{\tau}, \boldsymbol{c})=\{\boldsymbol{h}(\tau_1),..., \boldsymbol{h}(\tau_i), \boldsymbol{e}(x_1),..., \boldsymbol{e}(x_n), \boldsymbol{h}(\tau_{i+1}),..., \boldsymbol{h}(\tau_m)\}.$ (3) Finally, we can feed the embedding sequence to an LLM, and optimize the continuous prompts $\boldsymbol{\theta}^{\tau}$ using the downstream loss function $\mathcal{L}$ as follows: $${\boldsymbol{\theta}^{\tau}}^\star =\mathop{arg\ max}\limits_{\boldsymbol{\theta}^{\tau}}\ \mathcal{L}(f_{\Theta}(T_e^{'}(\boldsymbol{\tau}, \boldsymbol{c})), \boldsymbol{y}),$$ where $f_{\Theta}(\cdot)$ is the LLM function parametrized by $\Theta$. For a cloze-style input reformulated from general tasks, for example, the sentiment analysis task for the sentence ``I like the movie!" can be rephrased as a cloze-completion problem: ``I like the movie! It was [MASK].". The predicted words at the masked position are then employed for subsequent classification. In this case, a unique token [MASK] is integrated during the generation of the template in step (1), and a verbalizer $\phi$ is required to map class labels to words in the language model's vocabulary, e.g., \textit{positive}$\rightarrow$\textit{`great'}, resulting in: $${\boldsymbol{\theta}^{\tau}}^{\star} =\mathop{arg\ max}\limits_{\boldsymbol{\theta}^{\tau}} \sum_{\boldsymbol{c}}{\log{P\left([MASK]=\phi(\boldsymbol{y})|T_e^{'}(\boldsymbol{\tau}, \boldsymbol{c})\right)}}$$

The information condensed by the prompt falls into two categories: (1) \textbf{task-dependent prompt tuning}, and (2) \textbf{instance-dependent prompt tuning}. Each category encompasses general and specific enhancements for domain and task adaptation. Although some studies are based on PLMs, the advantages apply to LLMs, given the correlation between prompt tuning enhancements and model size \cite{lester-etal-2021-power} and successful implementations on large-scale PLMs. Moreover, it provides a parameter-efficient, fully controllable tuning method to adapt PLMs for more customized purposes.

\subsubsection{Task-dependent Prompt Tuning}
Task-dependent prompt tuning optimizes a shared prompt for all instances within a specific task, enabling it to encapsulate information from extensive datasets comprising thousands or millions of examples. However, training a naïve prompt is hard to converge and suboptimal for different scenarios, leaving room for improvement for specific tasks and domains.

\paragraph{Prompt Content Enhancement}
We refer prompt content as the embedding values of continuous prompt, enhancements are developed in terms of task-specific initialization and prior knowledge transfer. Pilot works have validated that in contrast to many optimizers that begin with a random distribution applied in general ML tasks, the optimization process of soft prompt is significantly influenced by its initial value. For language models, word embeddings are pre-trained to be quite distinct. Consequently, a standard optimizer such as stochastic gradient descent (SGD) can only update the parameters in a limited vicinity, leading to the possibility of falling into a local minimum \cite{allen2019convergence}. Therefore, a more effective initialization approach would involve using embeddings of concrete task-specific words. 

One of the pioneering works, WARP \cite{hambardzumyan2021warp} initializes the prompt by the embedding of special token ``[MASK]''. KnowPrompt \cite{chen2022knowprompt} designed learnable prompts as virtual type words and virtual answer words, which are initialized by the aggregated representation of concrete label words and disassembling words based on their frequency in the dataset. In addition, random initialization has been proven to be the least efficient, especially for small model, while Prompt-tuning \cite{lester-etal-2021-power} presented no significant gap between initialization strategies when the model size grows to 11B, indicating that LLMs is robust for prompt's initialization values in general tasks. 

Further studies have revealed that retrained prompts on source domains can enhance performance in unseen target domains, illustrating the ability of prompt transfer \cite{vu2021spot}. 
SPoT \cite{vu2021spot} initialize the prompt with a single generic source prompt learnt from multiple sources tasks, and then fine-tune it on target task in a classic way \cite{lester-etal-2021-power}. PPT \cite{gu2021ppt} also pre-trains a prompt using self-supervised learning on extensive unlabeled corpora, which then serves as the initial prompt for the target task. Su et al. \cite{su2022transferability} demonstrated the transferability of continuous prompts in both cross-task as well as cross-model settings, and find that a well-initialized prompt can significantly accelerate training convergence. Furthermore, take the advantage of transferability, LFPT5 \cite{qin2021lfpt5} employed soft prompt for lifelong learning. It continuously trains the prompt that simultaneously learns to solve the current task and generate training samples of previous tasks to overcome the \textit{catastrophic forgetting}. Progressive prompts \cite{razdaibiedina2023progressive} introduce the prompt tuning into continuous learning. The prompt for current task is defined as the concatenation of prompts that are optimized on previous tasks and a tunable current prompt.

\paragraph{Prompt Construction Enhancement}
We refer to the prompt construction about the positioning and length of the prompt, and combinations of additional templates or discrete prompts. Continuous prompts can be simply prepended, appended, and inserted to the original input sentences without extra language phrases. The pioneering study, WARP \cite{hambardzumyan2021warp} adopted all three intersections with a ``[MASK]'' token for classification tasks. In Prefix-tuning \cite{li2021prefix}, tunable prompts are prepended to the sentence embedding and the activation of all attention blocks, capitalizing on the left-to-right nature of the autoregressive model: the prepended prompt can efficiently affect the subsequent words through attention. In addition, a recent work \cite{lester-etal-2021-power} prepend prompts at the input layer, achieving comparable results to fine-tuned models.

Template is widely used to leverage the adaptation performance \cite{schick2020exploiting}, for example, reformulating an NLP task (e.g., sentence classification) into the masked words prediction task that is employed during LM pre-training. Based on the predefined task-specific templates, soft prompts can be inserted and thus offers flexibility for conditional tuning. KnowPrompt \cite{chen2022knowprompt} designed a template appending to the input sentence with a ``[MASK]'' between subject and object for relation extraction and incorporates trainable prompts of ``virtual type words'' surrounding these two entities. Output embeddings of ``virtual type words'' are trained to align logically with the target relation at the masked position, conditioning the optimization with the information of entity type. 
KiPT \cite{Li2022KiPTKP} developed a knowledge extractor for event detection tasks, which identifies trigger words in sentences based on their semantic similarity to event concepts above a threshold. Identified trigger words as well as the corresponding event labels will then be prepended to a randomly-initialized soft prompt with the input sentence. KiPT also reformulates the sequence tagging tasks: trigger-identification and trigger classification, into the generative task by outputting structured event records. 

\subsubsection{Instance-dependent prompt tuning}
A shared task-dependent prompt is static against the change in input sentence, which ignores semantic differences as well as specific knowledge of individual instances, and thus be suboptimal in fine-grained objectives. Instance-dependent prompt tuning however conditionally generates prompts for individual instances, incorporating both contextual information and task instructions.

\paragraph{Prompt Content Enhancement}
Enhancement of prompt content for instance-dependent tuning focus on learning a joint and adaptive representation of tasks as well as instance context. IDPG \cite{wu2022idpg} proposed an additional two-layer perceptron as a prompt generator, down and up project the sentence embedding to the adaptive soft prompt. ATTEMPT \cite{asai2022attempt} first train multiple prompts on large-scale source tasks, and calculate an aggregated prompt base on a sentence-wise attention network, which will then be mixed with a newly initialized target task prompt as the final instance-dependent prompt. Jin et al., \cite{jin2022instance} assume that prompt tokens differently contribute the instance, and thus designed a look-up module to score the association of prompt tokens to instance tokens, which is then used to calculate the aggregated prompt embeddings. Bhardwaj et al., \cite{Bhardwaj2022VectorQuantizedIS} generate context-aware prompts by a transformer-based sentence encoder, but further quantize the contextual prompt into a more compact representation to avoid optimization collapse. Levine et al., \cite{Levine2022StandingOT} learn the joint representation of prompt and input by a frozen T5 encoder following cross- and self-attention layers. Liu et al., \cite{Liu2022LatePT} propose an instance-aware prompt that is applied to the intermediate layers of LM. The proposed prompt generator is a simple feed-forward layer with bottleneck architecture which take the embedding of [CLS] token or pooling of embeddings of sentence tokens. 

\paragraph{Prompt Construction Enhancement}
Similar to the construction enhancement, instance-dependent prompt tuning introduce instance-dependent knowledge as concrete words or learn adaptive prompt in terms of positioning and length. OntoPrompt \cite{Ye2022OntologyenhancedPF} enriches the template with instance-related knowledge from external ontology as an additional text, and tune continuous prompts surrounding the ``[MASK]" to help prediction. Recently, to give a comprehensive discussion of the effect of content and structure of prompts, dynamic prompting \cite{Yang2023DynamicPA} proposed a unified framework to learn an instance-dependent prompt by dynamically defining prompt position, length, and values for each instance. It also proves the effectiveness of post-fix prompt, given most prior works prepend the prompt to the input sentence. 

\subsubsection{Open Challenges}
Continuous prompt tuning presents a streamlined method to utilize the broad language understanding capacity of LLMs for specific tasks across different domains. It efficiently tackles issues inherent in discrete prompt methods, such as (1) significant reliance on the prompt for LLM performance, where minor wording or template changes can greatly affect the result, (2) computational complexity in identifying the optimal natural language-based prompt from a large search space, and (3) the time-consuming and labor-intensive process of manually designing instructions, particularly in expertise-required domains. However, continuous prompt tuning has its limitations.
\begin{enumerate}
\item \emph{Interpretability} is often criticized as a weakness of soft prompt tuning. By discreting the optimal continuous prompts into nearby token vectors in LM's vacabulary, studies such as WARP \cite{hambardzumyan2021warp} have found these prompts to be non-interpretable and lacking meaningful content. In KnowPrompt \cite{chen2022knowprompt} and Prompt-tuning \cite{lester-etal-2021-power}, prompt tokens are discovered in close proximity to domain-related terms. For example, Prompts trained on the BoolQ dataset revealed that \textit{science}, \textit{technology}, and \textit{engineering} were the nearest neighbors of the optimal prompt, as approximately $20\%$ of the questions pertain to the ``Nature/Science'' category \cite{lester-etal-2021-power}. However, the interpretability of continuous prompts as a coherent sequence is still unclear. In addition, continuous prompt is not confined to directing LLMs using compact textual information. OPTIMA \cite{guo2022improving} achieves domain adapting with prompt by tuning it help regularizes the decision boundary to be smooth around regions where source and target data distributions are similar with an adversarial learning framework.
\item \emph{Limited access} to LLMs poses a significant challenge for continuous prompt learning, especially as models with immense sizes (e.g., 540B PaLM) and models with only API access. This restriction hinders differential optimization on continuous embeddings. In this case, derivative-free prompt tuning that optimizes the soft prompt without gradients from LLMs is widely discussed. Black-box tuning (BBT) \cite{Sun2022BlackBoxTF} proposed a gradient-free approach to searching the optimal prompt in a smaller intrinsic space.  with Covariance Matrix Adaptation Evolution Strategy (CMA-ES) for non-convex optimization instead of Adam. Similarly, Clip-Tuning \cite{Chai2022ClipTuningTD} use multiple deterministic clipping instances of the target LM to optimize an agent that learns the intrinsic prompt embedding. However, those methods still need access at least to the embedding layer, which is unsatisfactory for LLMs where only textual query is allowed. In this case, derivative-free approaches for discrete prompt search appear to be a more promising direction, and several studies have already achieved preliminary success \cite{deng2022rlprompt, prasad2022grips}.

\end{enumerate}

\section{Model Fine-tuning for Domain Specialization}\label{sec: knowledge_update}
\begin{wrapfigure}[10]{r}{9cm}
\vspace{-20pt}
\begin{center}
\includegraphics[width=0.48\textwidth]{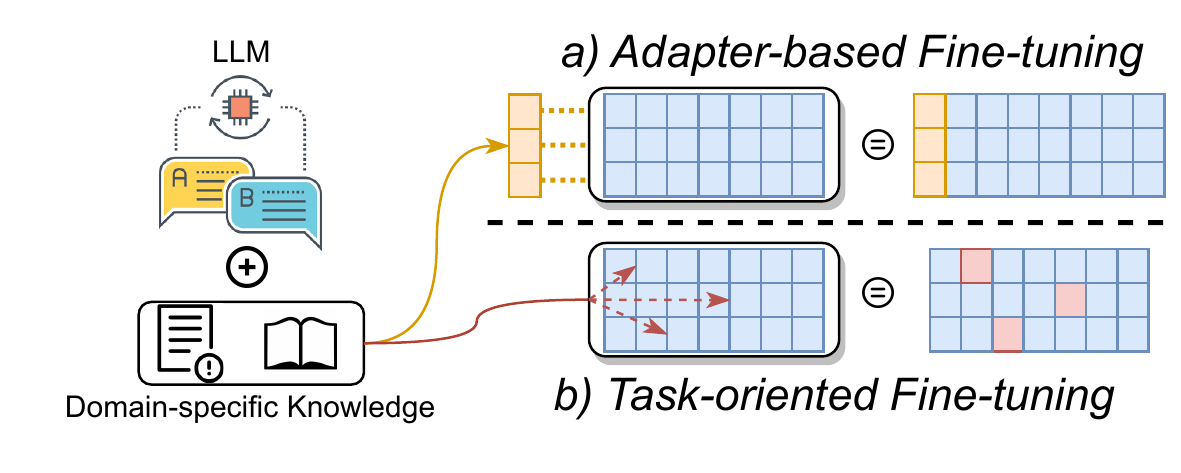}
  \end{center}
  \vspace{-6mm}
  \caption{The visualization of two approaches to fine-tune LLMs based on domain-specific knowledge, where the blue rectangle denotes the set of parameters in LLM. (a) the adapter-based LLM fine-tuning aims to fine-tune LLMs on specific domains with a small number of extra parameters (i.e., adapter); and (b) the task-oriented model fine-tuning aims to fine-tune LLMs based on specific tasks.}
  \vspace{-6mm}
  \label{fig: knowledge_update}
\end{wrapfigure}

LLMs, despite being trained on extensive general text data, might not encode adequate knowledge for specific tasks or domains. In such scenarios, fine-tuning the model on a smaller, domain-specific dataset can enhance its performance within that particular area. This fine-tuning can be divided into two main approaches. Adapter-based Fine-tuning and Task-oriented Fine-tuning. (1) \textbf{Adapter-based Fine-tuning:} This approach, as illustrated in Figure \ref{fig: knowledge_update} (a), employs neural adapters or modular components to enhance the LLM's performance on domain-specific tasks without major modifications to the LLM's inner parameters. These adapters, typically integrated into the existing LLM architecture, allow for task-specific learning while keeping the original model largely intact. (2) \textbf{Task-oriented Fine-tuning:} As represented in Figure \ref{fig: knowledge_update} (b), this method focuses on modifying the LLM's inner parameters to improve alignment with specific tasks. However, entirely updating all parameters of an LLM may be impractical due to hardware limitations and potential performance degradation. Therefore, the challenge for researchers lies in identifying which parameters require alteration within the expansive parameter space, or in efficiently updating a subset of these parameters. These two approaches allow LLMs to be tailored to specific tasks or domains, offering flexibility and efficiency in handling specialized applications.


\subsection{Adapter-based Fine-tuning}
Adapter-based fine-tuning aims to add a small number of extra parameters to the LLM for achieving better performance in specific tasks. Typically the additional parameters are encoded in simple modules to \textit{guide} the language model's adaptation to target domains or tasks. The golden spots for the added modules include: (1) simple with a small number of parameters; (2) extensible to the original language models; (3) flexible with sequential training on each specific domain. Most of the proposed strategies with the above favorable properties are built on adapters~\cite{houlsby2019parameter, rebuffi2017learning} under the umbrella of parameter-efficient fine-tuning. 

\subsubsection{Adapters}
Adapters are trainable modules inserted between layers of a pre-trained model~\cite{houlsby2019parameter}. The key property of adapters highlights that the parameters of the original language model keep frozen, thus provide sustainable parameter sharing even with varying domains and tasks. Suppose $f_{\Theta}(\cdot)$ denotes the function of LLM parametrized with the set of parameters $\Theta$ and $g_{\Delta \Theta}(\cdot)$ denotes the function of adapters with parameter $\Delta \Theta$, then $f_\Theta \circ g_{\Delta \Theta}$ represents the fine-tuned language model by adapters. Let $X$ be general input data with task performance metric $\phi$, and $D$ be the domain training data with domain-specific task performance $\phi_D$ (for both $\phi$ and $\phi_D$, a higher value indicates better performance), the goal of adapters is to find $g_{\Delta \Theta}$ such that:
\begin{align*}
    \phi(f_\Theta(X)) \approx \phi(f_\Theta \circ g_{\Delta \Theta}(X)) \quad\quad
    \phi_D(f_\Theta(D)) \leq \phi_D(f_\Theta \circ g_{\Delta \Theta}(D))
\end{align*}

Despite most empirical studies on cross-lingual or multi-task learning, some recent works explore unsupervised domain adaptation particularly using adapters. Unsupervised Domain Adaptation (UDA) using adapters has been explored in recent work, aiming to enhance the cross-lingual or multi-task learning capabilities of pre-trained models. The first attempt \cite{zhang2021unsupervised} targeted multi-domain adaptation with a two-step strategy: domain-fusion training with Masked Language Model (MLM) loss on a mixed corpus, followed by task fine-tuning with a task-specific loss on the domain corpus. Subsequently, UDApter was introduced, which also adopted the two-step training and fine-tuning approach, but segregated this into two adapter modules: a domain adapter and a task adapter. The domain adapter first learned domain-invariant representations, which were then concatenated with the task adapter whose parameters were frozen \cite{malik2023udapter}. This was achieved using the architecture defined in AdapterFusion \cite{pfeiffer2020adapterfusion}. AdapterSoup further improved adaptation efficiency by adopting a weight-average of domain adapters only during the testing phase \cite{chronopoulou2023adaptersoup}. To select domain adapters, three strategies were explored: exhaustive combination, text clustering, and semantic similarity.

Though these works focused on domain specialization, they were evaluated on pre-trained language models like GPT-2 \cite{zhang2021unsupervised, malik2023udapter, chronopoulou2023adaptersoup}, indicating potential applicability to larger language models. To address this, LLaMA-adapter was designed for efficient adaptation on Large Language Models with Adapters (LLaMAs) using self-instruct demonstrations. The adapter architecture incorporated a zero-init attention mechanism, and the domain specialization capability was tested on instruction-following and multi-modal reasoning tasks \cite{zhang2023llama}.

As the application of adapters expands, several techniques, while not explicitly claimed as effective for domain specialization, have either demonstrated potential by offering favorable performance on downstream tasks or served as integrated components in existing frameworks for domain specialization. Hence, adapters are usually classified based on their architectures into neural adapters and low-rank adapters. With the objective of facilitating user-friendly implementation, a growing body of work is dedicated to building comprehensive frameworks of different adapters \cite{pfeiffer2020adapterhub,hu2023llm}. Certain studies have also shown that adapter integration can yield superior performance across a variety of downstream tasks.

\paragraph{Neural adapters}

We call adapters with neural network architectures neural adapters. In their original design, ~\cite{houlsby2019parameter} uses a composition of down-projection, GeLU non-linearity~\cite{hendrycks2016gaussian} and up-projection with the feed-forward layers as the backbone. Later \cite{bapna2019simple} simplifies the architecture to a single hidden-layer feed-forward network and demonstrates the effectiveness on domain adaptation. The adapter modules are inserted after the multi-head attention and feed-forward layers in the transformer. These adapters have been named as bottleneck adapters or serial adapters. We use the latter throughout this paper when referring to \cite{houlsby2019parameter}.

The development of neural adapters naturally takes inspiration from neural network architecture design, such as ResNet, autoencoder, attention mechanism, etc. The adapters used in \cite{pfeiffer2020adapterfusion} have an additional residual connection. Soon after, \cite{pfeiffer2020mad} proposes MAD-X framework with invertible adapters, which are inserted adjacent to input and inverted to be fed into the output embeddings. At the high-level, invertible adapters can be considered a mimic of autoencoders. Tiny-attention adapter~\cite{zhao2022tiny} explores the effectiveness of adapters using attention with tiny per-head dimensionality. Till now, most proposed architectures apply fully-connected layers for down-projection and up-projection in adapters. However, Compacters~\cite{karimi2021compacter} considers parameterized hypercomplex multiplication layers~\cite{zhang2021beyond} as an alternative, which has a similar form as a fully-connected layer, but learns a sum of Kronecker products. The main advantage is parameter efficiency. Another way of achieving this is inspired by network pruning, as proposed by SparseAdapter\cite{he2022sparseadapter} to further reduce the training parameters by pruning at initialization. Note that SparseAdapter is a generic technique applicable to neural adapters. Congregating adapters via insertion can be considered as adaptation \emph{inside} the language models, an alternative is adaptation \emph{outside} the language models. $K$-adapters~\cite{wang2020k} proposes to train multiple adapters individually on various knowledge domains, then inject the learned knowledge with language models by concatenation. Recently, Sung et al.~\cite{sung2022lst} raises a concern on the high training memory required because the backpropagation flows throug the language model with inserted adapters in entirety. They further propose ladder side-tuning, which only adds small modules on the side of the language model connected to the language model backbone via shortcuts. Both techniques use MLP for demonstration, but keep flexible with different adapter architectures.

\paragraph{Low-rank adapters}

Low-rank adaptation (LoRA)~\cite{hu2021lora} is inspired by the observation that large language models reside on an intrinsic subspace~\cite{aghajanyan2020intrinsic}, where model parameters are efficiently updated. Therefore, learning in this subspace significantly reduces the amount of parameters. LoRA modules implant learnable SVD blocks as the subspace with a low matrix rank $r \ll d$, where $d$ is the dimension of input data. The matrices are added in parallel to the pre-trained weights, thus keeping them frozen during the fine-tuning. Notably, LoRA shows superiority in further reducing the number of trained parameters and introducing no latency during inference. 

A follow-up work on this line is DyLora~\cite{valipour2022dylora}, which addresses two issues of LoRA using dynamic search: fixed block size and exhaustive search on the optimal rank. Recently, another concern of LoRA was raised that low-rank modules have limited representation power, and further resolved by the Kronecker adapter (KronA)~\cite{edalati2022krona}. The essence is to substitute the SVD modules with a Kronecker product module with two matrices of smaller sizes. Despite not many follow-ups on the low-rank adapters, LoRA modules are included in various integrated adaptation frameworks~\cite{mao2021unipelt, he2021towards, wang2022adamix, hu2023llm} as an important building block. More details on these frameworks follow below.

\paragraph{Integrated adapter framework}
With the flourishing results on effective adapters as introduced above, it is a natural extension to incorporate several adapters of various families to boost their performance. AdapterFusion~\cite{pfeiffer2020adapterfusion} employs a straightforward idea: train multiple adapters on different tasks and combine the learned embeddings from each adapter with a fusion layer. UniPELT \cite{mao2021unipelt} proposes to activate different combinations of methods that best suit the current data or task setup via a gating mechanism. The sub-modules included serial adapter~\cite{houlsby2019parameter}, LoRA~\cite{hu2021lora}, Prefix-tuning~\cite{li2021prefix} and Bitfit~\cite{zaken2021bitfit}. Orthogonal to UniPELT, AdaMix~\cite{wang2022adamix} stacks multiple adapters of the same type, but avoids more computational cost by training the activation with stochastic routing. AdaMix can be regarded as a general technique that applies to any adapter, despite their implementation on only serial adapters and LoRA.

The idea of learning a routing function on an inventory of adapters further inspires follow-up works. In the context of multi-task learning, Polytropon~\cite{ponti2022combining} jointly learns an inventory of adapters and a routing function to re-combine the fine-tuned adapters of various sizes shared among different tasks. Variants of this scheme are further studied by~\cite{caccia2022multi}, including the replacement of the routing function with weights averaging, or multi-head routing function to achieve better expressivity. On the implementation-oriented aspect, AdapterHub~\cite{pfeiffer2020adapterhub} is the most comprehensive and easy-to-use library integrating all mainstream adapters. The only downside, however, is the absence of support of large language models. Recently, LLM-adapters~\cite{hu2023llm} introduces a framework including open-access large language models such as LLaMA, OPT, GPT-J, etc. It subsumes four adapters as basic components (Serial adapter~\cite{houlsby2019parameter}, MAD-X~\cite{pfeiffer2020mad}, Parallel adapter~\cite{yang2022unified} and LoRA~\cite{hu2021lora}) and remains extensible to new modules. The study of domain specialization further explores mathematical reasoning.

\subsubsection{Open Challenges}
The adapter's wide applications stem from its modular compatibility with language models, flexible design for integration, and efficient domain-specific data fine-tuning, advancing the adapter-based fine-tuning paradigm. However, these methods have drawbacks. Firstly, the performance of inserted modules can be sensitive to architectural design and size across different tasks and domains, risking insufficient representational power or overfitting on limited data. Secondly, additional modules enlarge the model size, imposing new resource demands and possibly extending inference time. Lastly, as Sung et al. note, the training memory needed by adapter-based methods remains substantial as backpropagation involves the entire model even when previous parameters are frozen \cite{sung2022lst}. Given these discussions, we outline the open challenges in applying adapters to LLMs for domain specialization:
\begin{enumerate}
    \item \textit{Stability and universality: } The performance of adapters can be subject to various architecture or hyper-parameters applied even on pre-trained language models (PTLMs), thus imposes a question mark on the stability and universality. This concern further extends to LLMs. A deeper understanding on how different adapters match with different task settings would be a significant boost to broader applications of adapters.
    \item \textit{Computational resources: } Adapters have shown remarkable results with a million-size of parameters on PTLMs. However, it remains unproven if they are enough for LLMs. If more adapter modules (more parameters) are required, then the issue of computational cost can be raised again. Another ideal spot on this issue is to reduce the training memory with novel architecture design or fine-tuning strategy. 
\end{enumerate}

\subsection{Task-oriented Fine-tuning}
Despite these incredible capabilities of LLMs trained on large text corpus, fundamentally improving the model performances beyond few-shot examples and auxiliary adapters still requires updating the inner parameters of LLMs on an extensive amount of high-quality domain-specific datasets. However, fine-tuning an LLM on any (domain) specific tasks poses two challenges: 1) updating LLM's global knowledge may destroy the in-context learning ability due to reasons including but not limited to overfitting, catastrophic forgetting, and task-specific biases \cite{wang2022preserving}. 2) fine-tuning LLMs is computationally expensive due to the vast parameter space and the deep model architecture. In this section, we review recent techniques on how to update the global knowledge of LLMs, which can be primarily categorized into two areas: \textbf{Instruction-based Fine-tuning} and \textbf{Partial Knowledge Update} to address both challenges, respectively.


\subsubsection{Instruction-based Knowledge Update} \label{sec:instruction_finetune}
\begin{figure*}[!t]
    \centering
    \includegraphics[width=\textwidth]{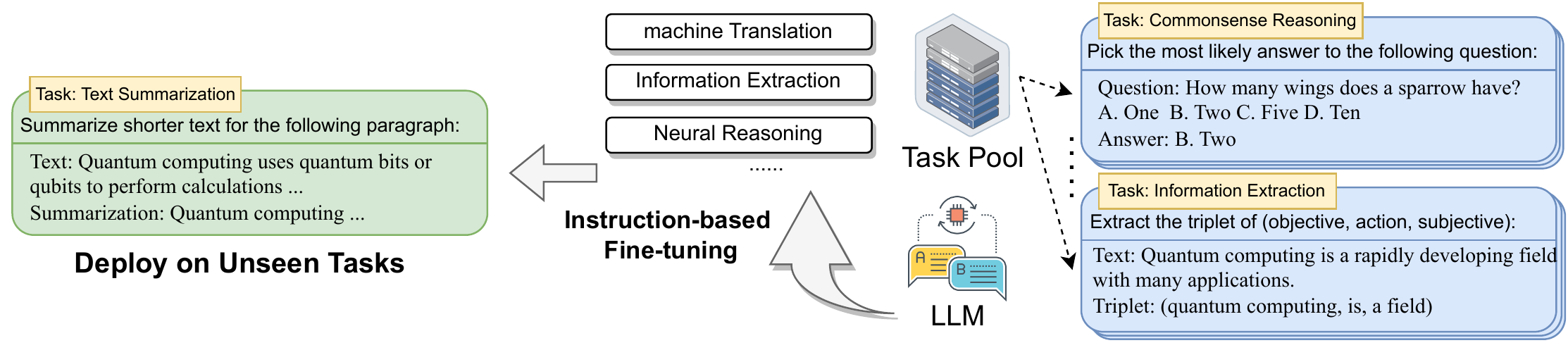}
    \vspace{-7mm}
    \caption{The overview of fine-tuning an LLM with explicit instructions across various domains and datasets. Particularly, the LLM is fine-tuned on a collection of tasks (e.g., commonsense reasoning, information extraction, etc.) with detailed instructions, and the fine-tuned LLM is expected to obtain problem-solving skills.}
    \vspace{-4mm}
    \label{fig: instruct_finetune}
    \end{figure*}

Instruction-based Knowledge Update refers to updating an LLM's parametric knowledge by fine-tuning LLMs on a diverse set of tasks with explicit instructions or prompts, which is conceptually the same as Instruct Learning introduced in \cite{ouyang2022InstructGPT}. An illustration of fine-tuning an LLM with instructions is provided in Figure \ref{fig: instruct_finetune}, where an LLM is fine-tuned on a collection of tasks across the whole NLP application domain, and the LLM is deployed on the held-out and unseen tasks. Wei et al. f\cite{wei2022finetuned} provided the very first attempt to fine-tune LLMs based on a collection of datasets described via instructions. Empirically, effective instructions can substantially improve zero-shot performance on unseen tasks. The instruction-tuned language model FLAN is fine-tuned on a 137B LLM over 60 NLP datasets using natural language instruction templates. The study shows that FLAN outperforms its unmodified counterpart and even surpasses both zero-shot and few-shot 175B GPT-3 on most unseen tasks. Subsequently, in recent works by \cite{chung2022scaling,menick2022teaching,huang2023language}, explicit instructions have been employed to fine-tune LLMs, with emphasis placed on (1) expanding the number of tasks, (2) enlarging the model's size, and (3) fine-tuning on chain-of-thought data. As a result, the fine-tuned LLM attains state-of-the-art performance on numerous benchmarks in both zero/few-shot NLP tasks.


\paragraph{Fine-tuning with Human Instructions} Fine-tuning with human instructions aims to guide LLMs towards generating safer, truthful, less toxic content in line with user intentions. Most LLMs utilize autoregressive approaches, making the generated content largely influenced by the training corpus distribution and less controllable. Reinforcement learning from human feedback (RLHF) is a notable technique for aligning LLM content with human needs \cite{christiano2017deep}. In RLHF: 1) LLMs create multiple content options for a prompt, ranked by humans for quality, relevance, and desired output alignment; 2) an external reward model assigns scores to content based on rankings, capturing evaluator preferences; 3) model policy is updated using reinforcement learning techniques to maximize expected reward, fine-tuning the model to better align with human preferences; 4) this process of content generation, ranking, reward modeling, and policy optimization repeats in iterations, with the model continually learning from human feedback. Existing methods successfully apply RLHF to fine-tune LLMs on complex reasoning tasks using human instructions \cite{wang2023document,peng2023instruction}.

\paragraph{Potential Limitations of Instruction-based Knowledge Update} Knowledge updates based on explicit instructions tend to perform well on Natural Language Understanding tasks but are limited to simpler instructions and struggle with tasks diverging from evaluation sets. Improving adaptability to diverse tasks often incurs catastrophic forgetting. A crucial question is extending model knowledge and abilities without causing such forgetting. Recently, Huang et al. proposed a method that uses a pre-trained LLM to generate high-confidence, rationale-augmented answers for unlabeled questions, improving general reasoning without ground truth labels or explicit instructions \cite{huang2022large}. Additionally, Scialom et al. are expanding LLM knowledge and abilities without forgetting previous skills by fine-tuning LLMs across various tasks and introducing an approach to counter catastrophic forgetting with Continual Learning via Rehearsal \cite{shin2017continual, scialom2022fine}.

\subsubsection{Partial Knowledge Update}
Other than leveraging task-specific instructions to fine-tune LLMs, a number of approaches emerge to conduct LLM fine-tuning by updating/editing a part of LLM parameters that link to specific knowledge without leveraging external guidance. Suppose $f_{\Theta}(\cdot)$ denotes the function of LLM parametrized with the set of parameters $\Theta$ and $\theta\in \Theta$ is the single parameter in $\Theta$. Updating the inner knowledge of $f_{\Theta}(\cdot)$ based on a collection of training data $D$ is denoted as:
\begin{equation}\label{eq: masking}
    \Tilde \Theta = \Theta + \nabla f_{\Theta}(D) \odot T, \quad\quad T^{(i)}= 
\begin{cases}
    1, & \text{if } \theta^{(i)} \in \Theta_T\\
    0,              & \text{if } \theta^{(i)} \notin \Theta_T
\end{cases}
\end{equation}
where $T$ denotes the mask vector and $T^{(i)} \in T$ denote the $i$-th element of $T$. The mask controls the amount of LLM's inner knowledge to be updated in each fine-tuning iteration, where we use $\Theta_T\subseteq \Theta$ to denote the parameters that need to be updated in $\Theta$. In the conventional setting of fine-tuning pre-trained language models \cite{howard2018universal,ziegler2019fine,roberts2020much}, $|\Theta| = |\Theta_T|$. However, updating all of the parameters is computationally prohibited and resource-consuming in the context of LLM. Empirically, $|\Theta| \gg |\Theta_T|$, which refers to the modification of only a small number of parameters. Existing parameter-efficient knowledge update can be categorized into three streams: i.e., \textbf{Knowledge Editing} aims at directly locating and updating a small subset of parameters in an LLM; \textbf{Gradient Masking} aims at masking out the gradients of non-relative parameters during the fine-tuning; and \textbf{Knowledge Distillation} focuses on obtaining a child model with domain-specific knowledge from LLMs.

\paragraph{Knowledge Editing} Recent research has seen success in updating LLMs with new memories to replace outdated information or add specialized domain knowledge. For instance, improving the ability to update an outdated prediction like ``Boris Johnson is Prime Minister of the UK" can enhance an LLM's reliability and generalization. Various methods have been proposed to locate and edit an LLM's parametric knowledge \cite{de-cao-etal-2021-editing,dai2021knowledge,meng2022locating,meng2022mass,hernandez2023measuring}. De Cao et al. proposed a hyper-network trained to update LLM parameters with a single fact needing modification, avoiding fine-tuning to prevent performance degeneration \cite{de-cao-etal-2021-editing}. However, later works found that hyper-network-based editing falters as the LLM scales up, proposing retrieval-based methods to store edits in explicit memory and reason over them to adjust LLM predictions \cite{mitchell2021fast,mitchell2022memory}. Other methods focus on localizing and understanding LLM internal mechanisms. Notable works identify crucial neuron activations for LLM factual predictions through attention mechanisms and causal interventions, successfully updating domain facts \cite{dai2021knowledge,meng2022locating,meng2022mass}. A recent method is proposed learning a map from textual queries to fact encodings in an LLM’s internal representation, using these encodings as knowledge editors and probes \cite{hernandez2023measuring}.


\paragraph{Gradient Masking} Gradient masking is a technique used to selectively update specific parts of an LLM during the fine-tuning process. The main goal is to reduce computational overhead and potentially mitigate issues such as catastrophic forgetting or overfitting, particularly when adapting pre-trained models to smaller or specialized datasets. Gradient masking involves modifying the gradients during back-propagation by applying a masking function (Equation \eqref{eq: masking}). This function determines which parts of the model will be updated, effectively \textit{masking} the gradients for certain parameters and keeping them unchanged. The choice of parameters to mask can be based on various criteria, such as their relevance to the task, importance in the model, or contribution to the overall loss.

Earlier attempts \cite{jiang2019smart,zaken2021bitfit} have been made to efficiently fine-tune relatively small language models by utilizing various regularization techniques, their methods cannot easily adapt to fine-tuning LLMs. This is primarily due to the substantially larger amounts of data and computational resources required to train LLMs effectively, which can be several orders of magnitude more than what is needed for smaller language models. To add gradient masks to LLMs,  CHILD-TUNING \cite{xu2021raise} utilizes the downstream task data to detect the most task-related parameters as the child network and freezes the parameters in non-child
network to their pre-trained weights. Moreover, Zhang et al. \cite{zhang2022fine} propose a Dynamic Parameter Selection algorithm for efficiently fine-tuning LLMs, which adaptively selects a more promising sub-network to perform staging updates based on gradients of back-propagation, which brings great improvement in domain-specific downstream tasks under low-resource scenarios.

\paragraph{Knowledge Distillation} 
While most works on LLM self-knowledge update focus on task-specific instructions and parameter efficiency, a promising area of research explores distilling domain-specific knowledge from LLMs into smaller networks to reduce inference latency and enhance domain-specific task solving ability. Muhamed et al. compressed a 1.5 billion parameter LLM to a 70 million parameter model for Click-through-rate prediction, introducing twin-structured BERT-like encoders and a fusion layer for a cross-architecture distillation from a single LLM, resulting in superior performance in both online and offline settings \cite{muhamed2021ctr}. Similarly, \cite{vucetic2022efficient,marjieh2023language,azerbayev2022explicit} employ a knowledge distillation module for LLM fine-tuning, achieving faster convergence and better resource utilization. This module leverages pre-trained parameters for quick convergence and trains a small subset of parameters to address model over-parameterization. Furthermore, \cite{shridhar2022distilling,hsieh2023distilling} distill the step-by-step chain-of-thought reasoning abilities of larger models into smaller models.

\subsubsection{Open Challenges}
Fine-tuning LLMs with the latest data ensures that they provide relevant and accurate information, especially in domains where rapid changes occur, such as technology, medicine, and current events. Furthermore, we have observed different applications or users may have unique requirements or preferences. However, fine-tuning the large-scale LLMs also poses several open challenges:
\begin{enumerate}
    \item \textit{Compliance with regulations}: In most cases, updating and fine-tuning LLMs are necessary to ensure compliance with specific regulations or guidelines, such as data protection laws or industry-specific requirements. The so-called \textit{LLM alignment} can be accomplished during the fine-tuning phase.
    \item \textit{Computational resources}: Fine-tuning or updating inner knowledge of LLMs necessitates access to high-performance GPUs or specialized hardware, which can be expensive and difficult to obtain, particularly for individual researchers or smaller organizations. Pursuing fine-tuning efficiency is still a practical yet essential problem.
\end{enumerate}

\section{Applications of LLM Domain Specialization}\label{sec: application}
In this survey paper, we explore the applications of LLMs across a range of domain-specific tasks in social sciences (e.g., education, finance, law), natural sciences (e.g., biomedicine, earth science), and formal sciences (e.g., human-computer interaction, software engineering, and cyber security). To achieve domain specialization for LLMs in these diverse fields, readers can employ various techniques, such as external augmentation, instruction crafting, and knowledge update. These approaches can help tailor LLMs to specific tasks and challenges in each domain, enabling more accurate, relevant, and effective applications. Although each domain has its unique challenges and requirements, several common applications of specialized LLMs are shared across these fields:
\begin{itemize}
    \item \textit{Advanced information extraction}: They can identify entities, relationships, and events from domain-specific texts, such as recognizing genes in biomedical literature or detecting legal clauses in contracts.
    \item \textit{Text generation and summarization}: They can generate high-quality, domain-specific content and create accurate summaries of complex domain-specific texts.
    \item \textit{Data-driven predictions and recommendations}: They can analyze domain-specific data for forecasting and providing recommendations, like predicting financial trends or suggesting personalized medical treatment plans.
    \item \textit{Conversational agents and expert systems}: They can be incorporated into conversational agents or expert systems for domain-specific guidance, such as virtual tutors or legal chatbots.
    \item \textit{Automated code generation and analysis}: In software engineering, they can generate or analyze code, identify bugs, or suggest improvements based on natural language descriptions.
\end{itemize}
In this section, we dive deep to review existing techniques for specializing LLMs in domain-specific tasks and discuss potential open challenges in detail. Due to the space limitation, we only provide a brief introduction of each domain and leave the complete discussion in the supplementary material.

\paragraph{Biomedicine}
Language models are becoming increasingly useful in the field of biology, from fundamental biomedical research \cite{mahjour2023designing,ross2022large} to clinical healthcare support \cite{rao2023evaluating, jeblick2022chatgpt, moons2023chatgpt}. At the fundamental biomedicine science level, LLMs can be trained on vast amounts of domain-specializing data (e.g., genomic and proteomic) to analyze and predict biological functions, disease mechanisms, and drug discovery. LLMs can also aid in predicting protein structures and interactions, which are critical for understanding cellular processes and designing new drugs. At the clinical healthcare support level, pre-trained or medical corpus fine-tuned LLMs can be used for the natural language processing of medical records to identify patterns, make diagnoses, and provide personalized treatment recommendations. Also, LLMs can assist in medical image analysis in a multi-modality learning way, such as identifying specific features in X-rays or MRI scans. Overall, LLMs offer tremendous potential for advancing biology research and improving healthcare outcomes.

\paragraph{Earth Science}
Earth science is an interdisciplinary domain focused on examining the interactions between physical and human systems across diverse spatial and temporal scales. This field incorporates methods from Earth observation, information science, spatial analysis, complexity theory, and simulation modeling to investigate phenomena like climate change, land-use change, natural disasters, environmental development, and urbanization. Spatial information is vital to Earth science, and geographic information science tools are invaluable for a wide range of interdisciplinary studies involving spatial data. Large language models like ChatGPT can act as question-answer systems, assisting those interested in Earth Science to gain pertinent knowledge, such as recommending the optimal earth observation dataset for specific research purposes, offering code examples like Google Earth Engine code for processing satellite data, providing high-quality responses to environmental-related questions \cite{zhu2023chatgpt}, developing innovative idea \cite{biswas2023potential}, generating climate scenario \cite{biswas2023potential}. LLMs can also be tailored to various Earth Science-related downstream tasks through methods such as fine-tuning, few-shot, or even zero-shot learning. 

\paragraph{Finance and Law}
Specializing LLMs in the financial and legal domains requires careful adaptation to the distinctive characteristics of these fields. In the financial domain \cite{wu2023bloomberggpt,leippold2023sentiment,yang2023large,lopez2023can}, models need to comprehend complex financial terminologies, economic trends, and regulatory norms to accurately generate content like financial reports, investment analyses, or risk assessments. Meanwhile, the legal domain \cite{chalkidis2020legal,prasad2022effect,valvoda2023on} demands understanding and generation of intricate legal language, comprehension of laws, legal codes, and court rulings, while maintaining absolute precision and a xxformal tone. For both domains, model specialization often involves fine-tuning with domain-specific datasets, incorporating explicit domain knowledge, and optimizing for domain-specific objectives like compliance with regulations, accuracy of information, or effectiveness of advice. However, it's crucial to maintain an ethical guardrail for these models, given the high stakes nature of both financial and legal decisions. The specialized models also need to keep abreast of the evolving landscapes of these domains, adapting to changes in laws, regulations, or financial trends.

\paragraph{Human Computer Interaction and Software Engineering}
Specializing LLMs in the domains of human-computer interaction (HCI) and software engineering requires a deep understanding of the terminologies, workflows, and conventions unique to these areas. In the HCI domain, an LLM may be specialized to understand and respond to user inputs more effectively, potentially improving the design and usability of interfaces by offering more natural and intuitive interaction paradigms. This involves training the model on diverse data, ranging from human conversational data to user interaction logs. On the other hand, in the software engineering domain, an LLM can be specialized to aid in tasks such as code generation, bug detection, code review, and documentation. This involves training the model on large codebases, issue trackers, documentation, and other software-related data. These specialized models can provide valuable assistance to developers, enhance the software development process, and potentially revolutionize the way we interact with computers. Despite the promising applications, several challenges remain, including the complexity of these domains, the need for accurate and up-to-date data, and the balance between specialized and general knowledge.

\section{Open Challenges and Future Works}\label{sec: open_challenges}

\subsection{Open Challenges}
After having explored the current approaches for domain specialization of large language models, categorized as white box, grey box, and black box methods, it is essential to acknowledge that despite the significant progress in this field, there remain several open challenges. These challenges pervade all categories of models, irrespective of their accessibility or the specific techniques employed for specialization. As we strive to create LLMs that can effectively understand and generate domain-specific content, it is these challenges that will shape the future trajectory of research in this field. Let us delve into these open challenges to better comprehend the complexities of domain specialization and the areas where further research is required.
\begin{itemize}
    \item \textit{Domain Complexity}: Each domain has its unique intricacies and complexities, which could range from highly specialized vocabularies, and nuanced terminologies to complex knowledge structures. For instance, the legal or medical field employs language and terms that are extremely domain-specific and follow certain syntax and structure rules. This complexity extends to the relationships between different entities and concepts within the domain. Accurately understanding and modeling this intricate domain knowledge is a significant challenge for all types of models.
    \item \textit{Balancing General and Domain Knowledge}: An LLM, while needing to understand the specificities of a particular domain, also has to maintain its general knowledge to provide contextually appropriate responses. If a model is overly specialized, it may perform exceptionally within the targeted domain but fail to understand or generate coherent responses to prompts outside of it. Conversely, retaining too much general knowledge may dilute the domain-specific responses. Striking this balance between general and domain knowledge is a complex task.
    \item \textit{Explainability and Trust}: As LLMs become more sophisticated, their decision-making process also becomes more opaque, raising the challenge of explainability. It is crucial for users, especially in high-stakes domains like healthcare, law, or finance, to understand how the model arrived at a certain output. Achieving this transparency can help build trust in the system. The challenge lies in the trade-off between model complexity and explainability, as increasing one often decreases the other.
    \item \textit{Adapting to Domain Evolution}: Domains are not static; they evolve over time with the introduction of new terminologies, concepts, and trends. For example, the ongoing COVID-19 pandemic introduced a slew of new medical terms and concepts. Therefore, an LLM that is specialized for a certain domain must continuously adapt to these changes to stay relevant and effective. Designing models that can keep pace with the evolving landscape of their specialized domain is a challenging task.
    \item \textit{Scalability}: Domain specialization often involves training or fine-tuning the LLM with domain-specific data, crafting specific prompts, or using other domain-specific resources. While this might be feasible for a few domains, scaling this process to cover a wide range of domains or to handle large, complex domains is a significant challenge. It involves not just computational resources but also the availability of domain-specific data and expertise. The challenge is to create efficient and effective methods for domain specialization that can be scaled to cover many different domains.

\end{itemize}

\subsection{Future Works}
As we chart the frontier of large language model specialization, it's not only crucial to build upon and refine the existing black-box, grey-box, and white-box methods, but also to anticipate and explore innovative, out-of-the-box techniques that have the potential to transcend these conventional approaches. Leveraging the rapid advancement of AI technologies and the increased understanding of LLMs, we can envision a future where novel techniques will emerge that push the boundaries of what's possible in domain specialization, providing enhanced performance, greater flexibility, and more efficient utilization of resources.
\begin{itemize}
    \item \textit{Hybrid Approaches}: This could involve combining multiple methods depending on the stage or specific needs. For example, a model could start with a black-box approach, using external resources to augment input prompts, then proceed to a grey-box method where gradients or loss values are used to further refine the prompts, and finally employ a white-box approach to fine-tune the model based on the learned strategies and feedback. This hybrid approach could provide a balance between resource requirement and model performance and might be especially effective when dealing with scarce domain-specific data.
    \item \textit{Meta-Learning or AutoML Techniques}: AutoML or meta-learning strategies could be used to automate the process of selecting the best strategies for domain specialization. For instance, a meta-learning approach might learn a policy to select the best data for fine-tuning, the best prompt engineering techniques, or the best layers to fine-tune for a given domain, based on previous experience with similar domains. This could significantly reduce the resources and expertise needed for domain specialization, and could potentially lead to more effective and efficient methods.
    \item \textit{Incorporating More Explicit World Knowledge}: Instead of relying solely on text-based pre-training, future LLMs might leverage structured knowledge sources, like knowledge graphs, to augment their understanding of the domain. This could involve techniques like graph neural networks or attention mechanisms that operate on graph-structured data. For instance, a medical LLM could incorporate knowledge from a medical ontology graph to better understand the relationships between various medical terms and concepts. This could lead to more accurate and informative outputs, especially in domains where explicit structured knowledge is available.
    \item \textit{Human-in-the-loop Learning}: This involves continuous interaction and feedback from human users or experts to guide the model's learning process. For instance, a legal LLM could be continuously updated based on feedback from legal professionals using the model. This feedback could be incorporated in the form of additional training data, changes to the model's reward function in a reinforcement learning framework, or modifications to the model's prompts. This could lead to a more dynamic and adaptable model that can evolve with the needs and knowledge of the users.
    \item \textit{Active Learning}: This approach involves the model actively querying for information or feedback when it encounters a domain-specific concept it doesn't understand or has low confidence about. For instance, if a model trained on general news articles encounters a specialized medical term it doesn't understand, it could query a medical ontology or ask for clarification from a human user. The model could then incorporate this new information into its subsequent responses. This could make the model more effective at handling unfamiliar domain-specific topics, and could provide a more interactive and engaging user experience.
\end{itemize}

Each of these techniques provides a promising direction for future research in the domain specialization of large language models, and could help address some of the challenges and limitations of the current black-box, grey-box, and white-box methods.

\section{Conclusion}
In conclusion, the rapid advancement of LLMs has sparked significant interest in harnessing their potential to tackle domain-specific tasks in various natural, social, and formal science fields. However, several challenges, such as limited domain-specific expertise, knowledge elicitation, and model complexity, hinder the direct application of LLMs in these domains. This survey systematically categorizes and summarizes existing domain specialization techniques based on their access level to LLMs, along with a comprehensive overview of application domains that can benefit from specialized LLMs. By offering a detailed analysis of the advantages, disadvantages, and relationships among different techniques and domains, this survey aims to assist domain experts in identifying suitable techniques for their target problem settings, while also providing data scientists with a clear understanding of the practical significance and open challenges in various application domains. Moreover, the paper highlights the current status of research in this area, shedding light on future trends and potential avenues for interdisciplinary collaboration. As the field of LLM domain specialization continues to evolve, this survey serves as a valuable resource for researchers and practitioners, fostering further advancements and innovations in the application of artificial intelligence across diverse domains.

\bibliographystyle{ACM-Reference-Format}
\bibliography{main}


\end{document}